\documentclass{article}

\PassOptionsToPackage{numbers}{natbib}

\usepackage[final]{neurips_data_2021}

\usepackage[utf8]{inputenc} 
\usepackage[T1]{fontenc}    
\usepackage{hyperref}       
\usepackage{url}            
\usepackage{booktabs}       
\usepackage{amsfonts}       
\usepackage{nicefrac}       
\usepackage{microtype}      
\usepackage{xcolor}         

\usepackage{subcaption}
\usepackage{booktabs}

\usepackage{times}
\usepackage{epsfig}
\usepackage{graphicx}
\usepackage{amsmath}
\usepackage{amssymb}
\usepackage{wrapfig}
\usepackage{enumitem}

\usepackage{tikz}
\usetikzlibrary{trees}
\usetikzlibrary{bayesnet}
\usetikzlibrary{backgrounds}
\usepackage{algorithm}
\usepackage{algpseudocode}

\usepackage{marvosym}

\usepackage[weather]{ifsym}

\usepackage{wasysym}
\usepackage{fontawesome}

\title{A Procedural World Generation Framework for Systematic Evaluation of Continual Learning}

\author{Timm Hess, Martin Mundt, Iuliia Pliushch, Visvanathan Ramesh\\
	Goethe University, Frankfurt am Main, Germany\\
	{ \tt\small hess@ccc.cs.uni-frankfurt.de} \\ 
	{\tt\small \{mmundt, pliushch, vramesh\}@em.uni-frankfurt.de}  
}

\begin{document}
	
	\maketitle
	
	\begin{abstract}
		Several families of continual learning techniques have been proposed to alleviate catastrophic interference in deep neural network training on non-stationary data. However, a comprehensive comparison and analysis of limitations remains largely open due to the inaccessibility to suitable datasets. Empirical examination not only varies immensely between individual works, it further currently relies on contrived composition of benchmarks through subdivision and concatenation of various prevalent static vision datasets. In this work, our goal is to bridge this gap by introducing a computer graphics simulation framework that repeatedly renders only upcoming urban scene fragments in an endless real-time procedural world generation process. At its core lies a modular parametric generative model with adaptable generative factors. The latter can be used to flexibly compose data streams, which significantly facilitates a detailed analysis and allows for effortless investigation of various continual learning schemes. 
	\end{abstract}

	\section{Introduction} \label{section:introduction}
	In an era where deep neural networks have diffused into every conceivable application, a natural interest in the long-standing challenge of \textit{catastrophic interference} \cite{McCloskey1989, Ratcliff1990} in continuous training has resurfaced. Various families of approaches have emerged to alleviate this challenge of formerly encoded representations being rapidly superseded with the arrival of novel distinct data from a non iid data distribution continuum \cite{Ratcliff1990,Rebuffi2017,Isele2018,Rolnick2018, Robins1995,Shin2017,Mundt2020a, Li2016,Rannen2017,Zhai2019, French1992,Kirkpatrick2017, Zenke2017, Chaudhry2018}. However, despite the asserted progress, recent reviews repeatedly stress the importance of more exhaustive and realistic evaluation \cite{Pfulb2019,DeLange2019,Lesort2019, Diaz-Rodriguez2018, Kemker2018, Farquhar2018a, Lopez-Paz2017, Farquhar2018a, Mundt2020b}. 
	Notably in computer vision, the majority of presently emphasized benchmarks are contrived variants of the prevalent existing datasets \cite{LeCun1998,Nilsback2006,Krizhevsky2009,Xiao2010,Netzer2011,Russakovsky2015}, where individual concepts of the datasets are split into disjoint subsets and presented to the learner in sequence, or deliberately designed to follow this trend of object and class increments \cite{Lomonaco2017,Rebuffi2017a}. In hindsight, the latter benchmark construction neglects two imperative elements. First, it disregards a myriad of real-world continual learning scenarios, where the environment and its various conditions can be subject of constant change. Second, the uncontrolled data acquisition process hinders insights on a method`s feasibility beyond the specific empirical outcome. 
	
	We posit that generation of synthetic data through the use of parametrized generative models can provide a remedy for the present lack of a more detailed continual learning analysis. Here, the crucial realization is that catastrophic forgetting is a consequence of dense entangled representations in neural networks being greedily overwritten by newly encountered information. As such, the catastrophic interference phenomenon is a result of chosen optimization strategies and likewise applies to any investigation of synthetic data. More so, we conjecture that: \textit{if catastrophic forgetting cannot be circumvented in scenarios with a known synthetic data foundation, there is limited hope to understand limitations and overcome the challenge in real-world settings.} 
	
	In principle, the idea to leverage virtual data has already found countless prior applications, primarily due to simulators' ability to yield automatic precise ground truth information \cite{Li2012,Satkin2012,Vazquez2014,Peng2015,Hattori2015,Handa2016,Richter2016,Dosovitskiy2017,DeSouza2017,Jiang2017,Saleh2018}. Computer-graphics frameworks thus typically facilitate sampling of a maximum of conceivable variations to enable deep learning in domains where scares data is available and real-world data acquisition is insurmountable. However, in continual learning, scenarios of interest should inherently encompass knowledge about the detailed temporal shifts in the observed distribution. These range from occurrence of particular objects, their geometry and texture, the frequency and order of objects' (dis-)appearance in the scene, or continuous changes in the environmental weather and lighting. Corresponding investigations of continual learning thus require straightforward accessibility to meticulous control of the real-time online changes in the independent generative factors. With a focus set on large-scale annotated data generation to overcome an existing lack of data, the latter nuanced control is generally not exposed to the user in existing simulators' surface controls. 
	\begin{figure}[t]
		\begin{subfigure}{\textwidth}
			\centering
			\begin{tabular}{p{0.0025\textwidth}}
				\vspace*{-2.75em} \hspace*{-0.75em} \textbf{A}
			\end{tabular}
			\includegraphics[width=0.23\textwidth]{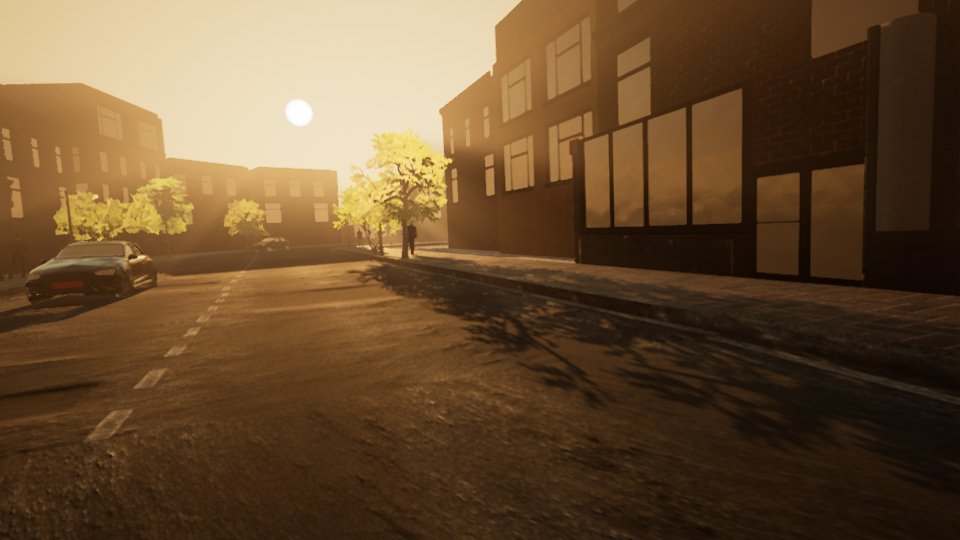} 
			\includegraphics[width=0.23\textwidth]{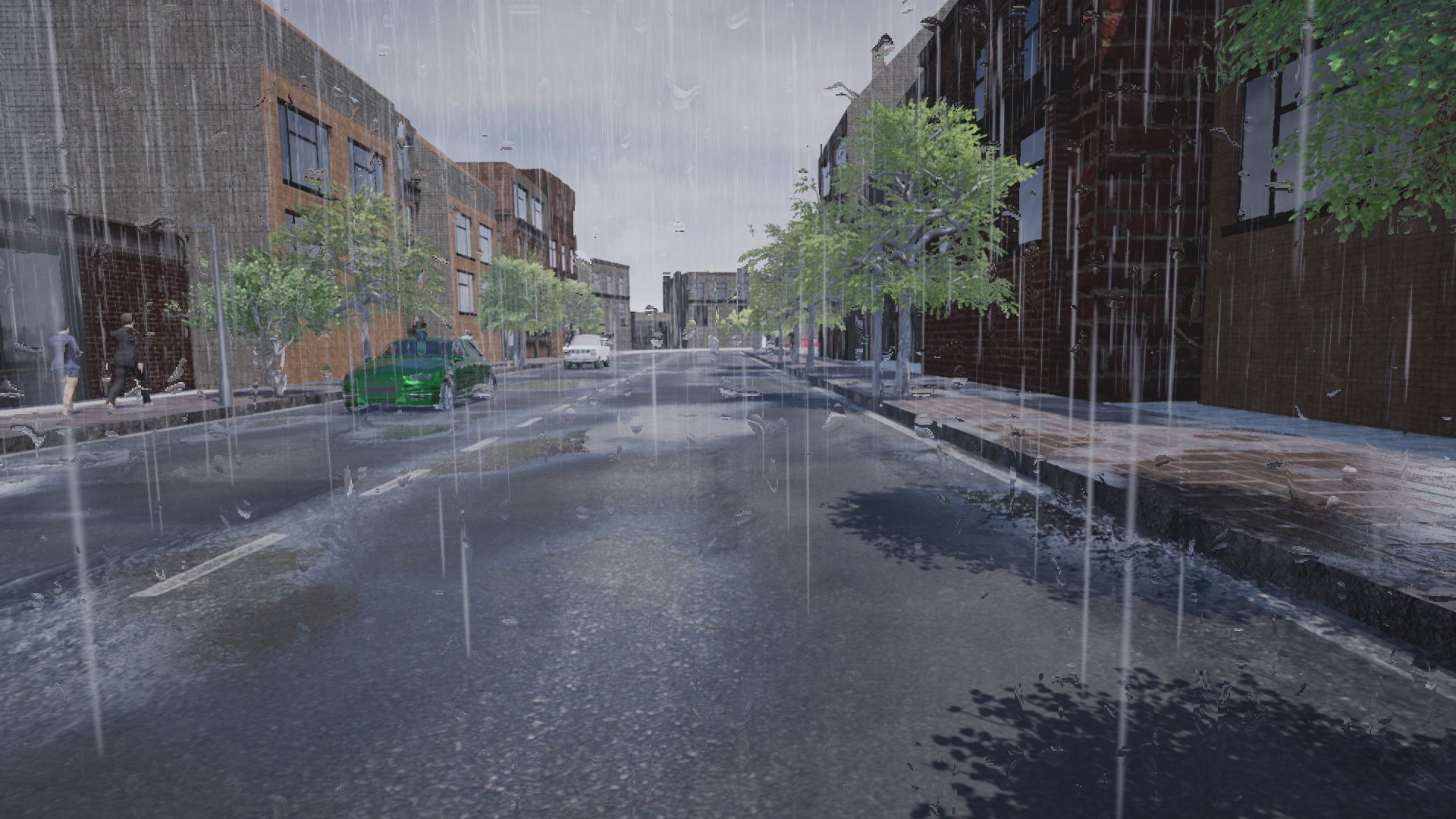}
			\hspace*{0.005\textwidth}
			\includegraphics[width=0.23\textwidth]{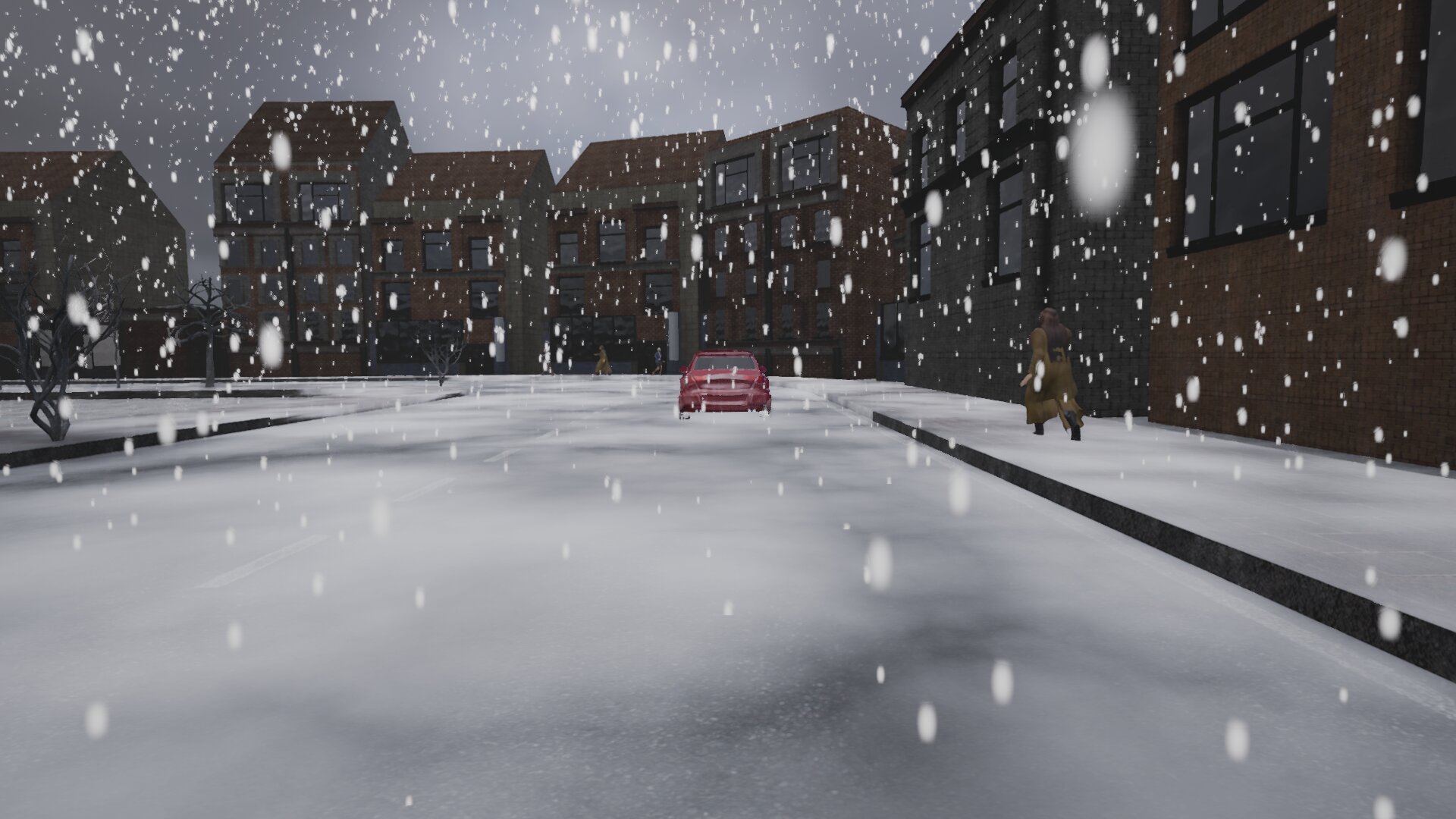} 
			\includegraphics[width=0.23\textwidth]{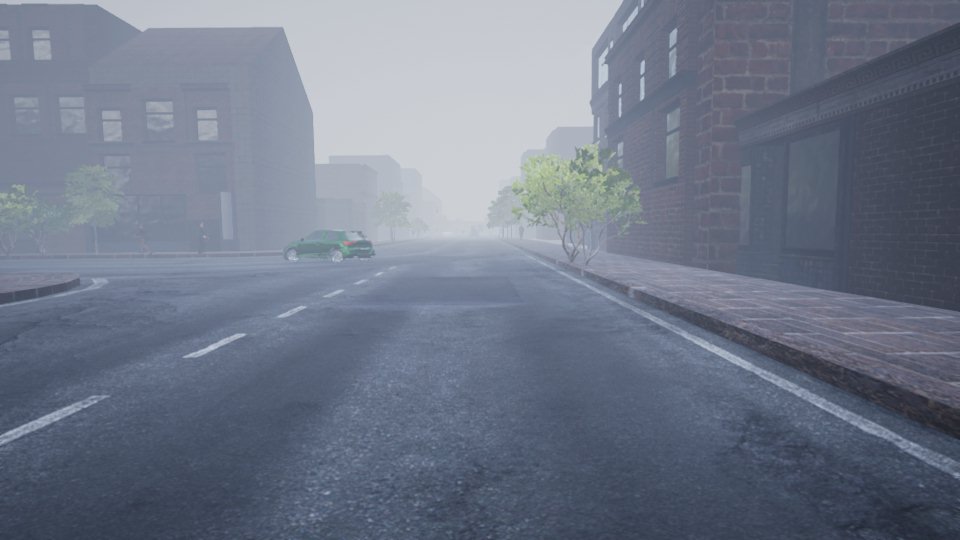}
		\end{subfigure}
		\par\smallskip
		\begin{subfigure}{\textwidth}
			\centering
			\begin{tabular}{p{0.0025\textwidth}}
				\vspace*{-2.75em} \hspace*{-0.75em} \textbf{B}
			\end{tabular}
			\includegraphics[width=0.23\textwidth]{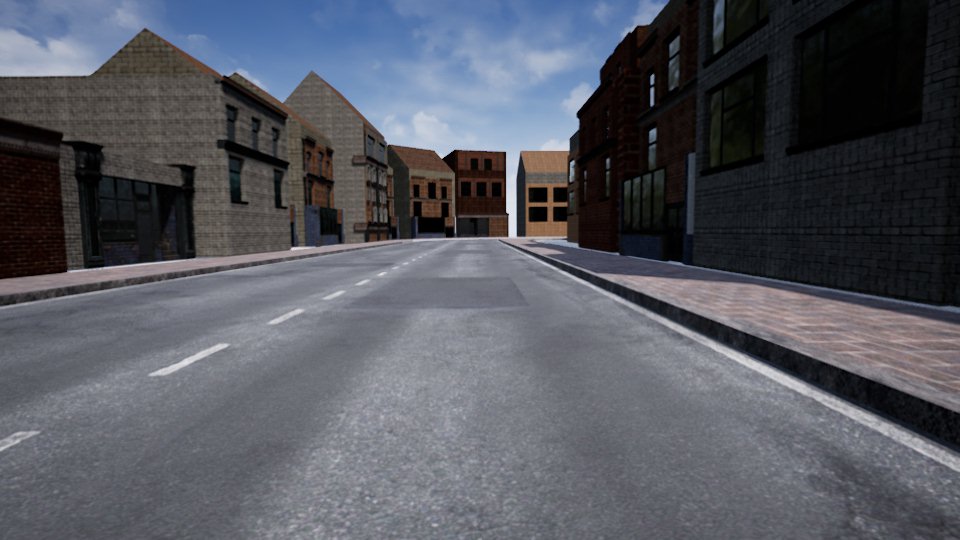} 
			\includegraphics[width=0.23\textwidth]{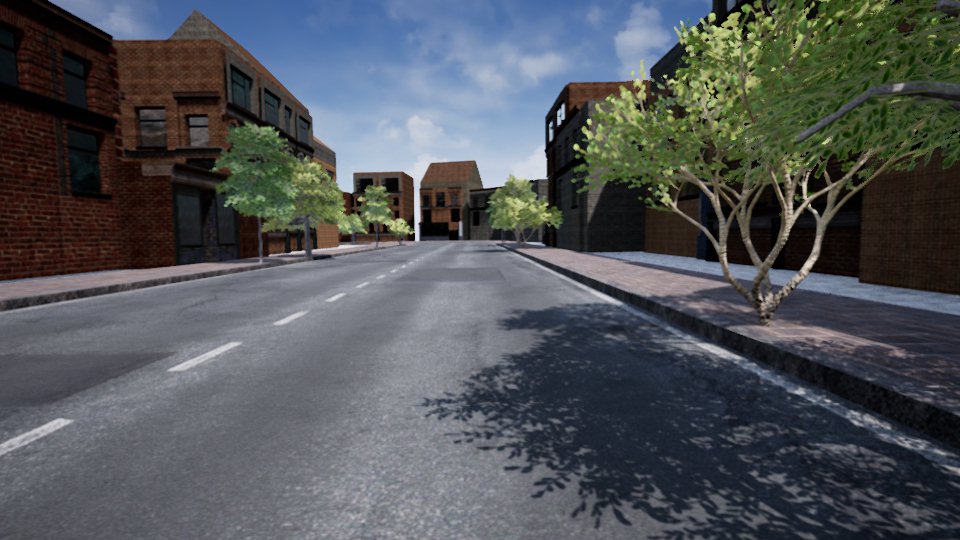}
			\hspace*{0.005\textwidth}
			\includegraphics[width=0.23\textwidth]{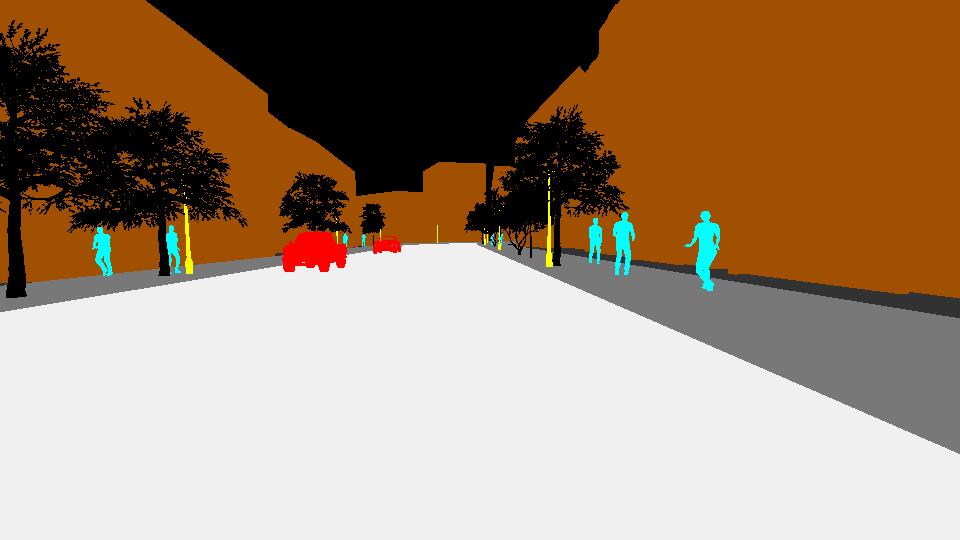}
			\includegraphics[width=0.23\textwidth]{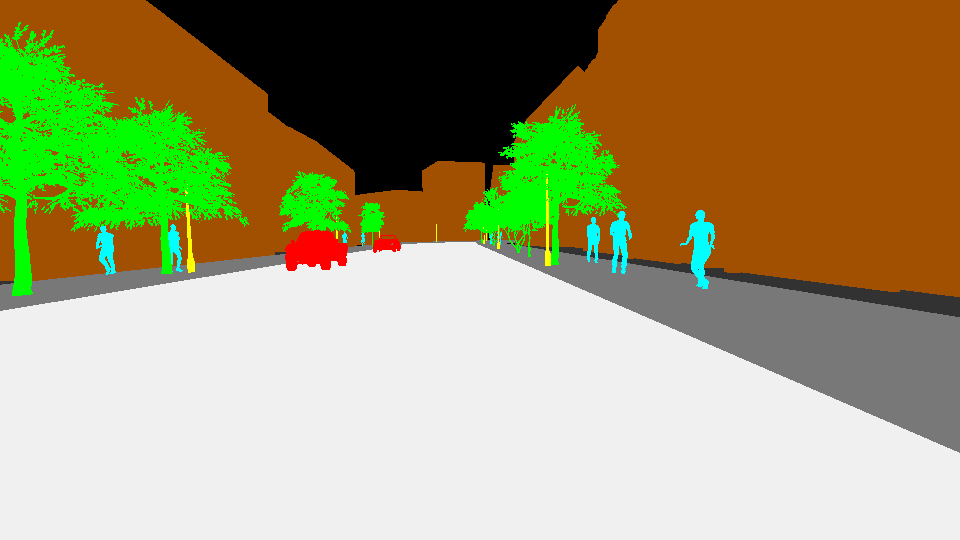}
		\end{subfigure}
		\par\smallskip
		\begin{subfigure}{\textwidth}
			\centering
			\begin{tabular}{p{0.0025\textwidth}}
				\vspace*{-2.75em} \hspace*{-0.75em} \textbf{C}
			\end{tabular}
			\includegraphics[width=0.23\textwidth]{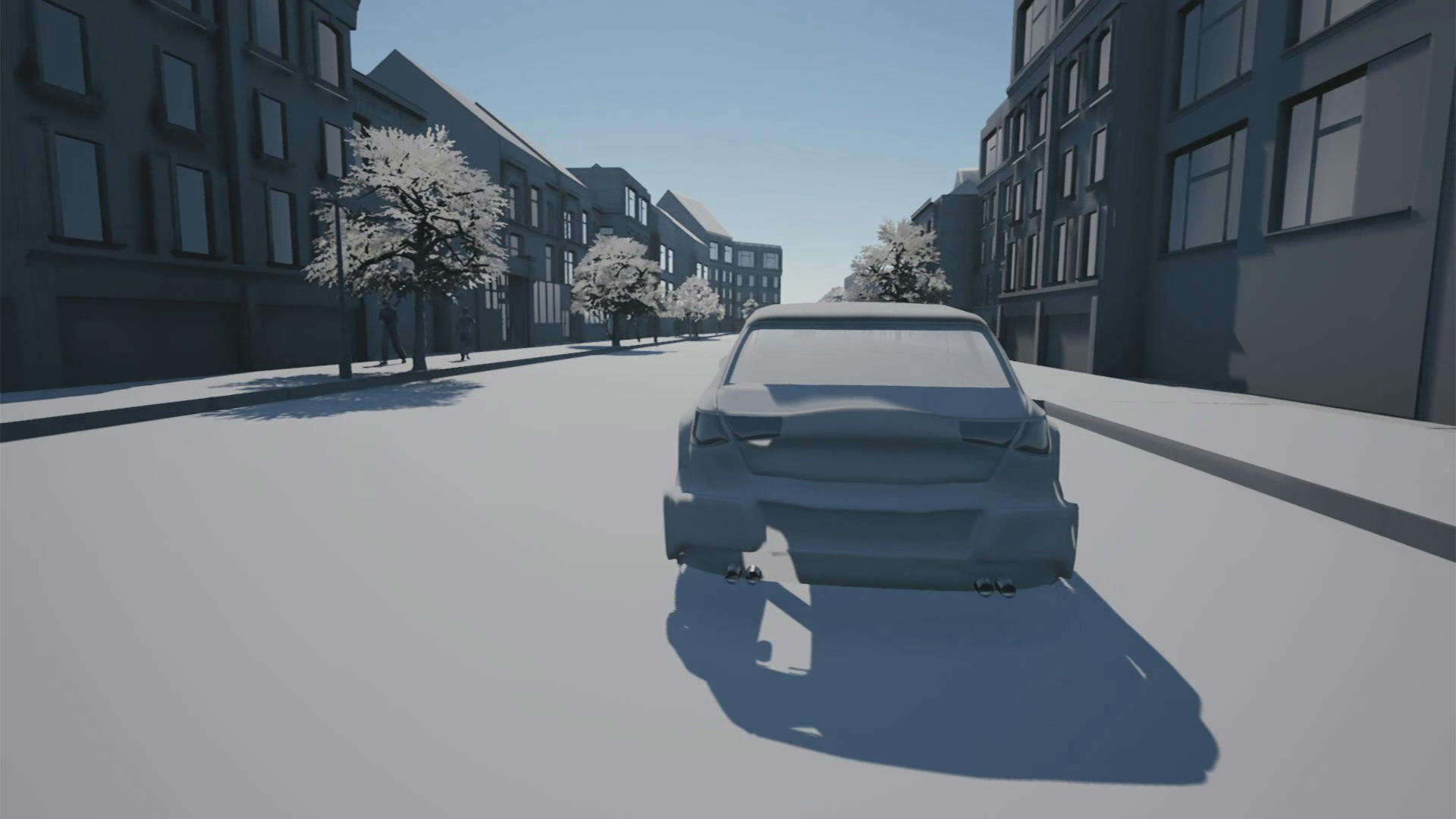} 
			\includegraphics[width=0.23\textwidth]{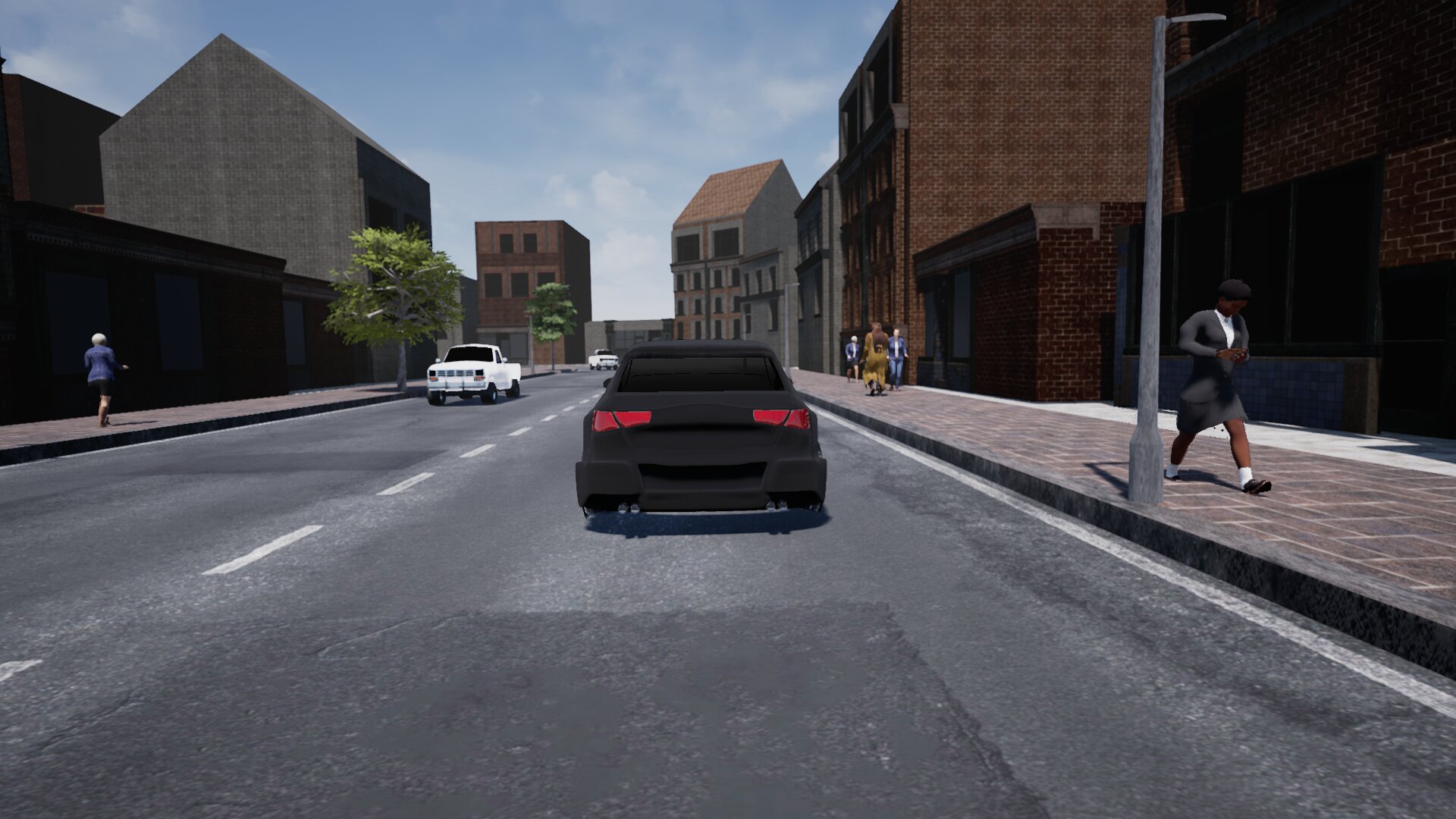}
			\hspace*{0.005\textwidth}
			\includegraphics[width=0.23\textwidth]{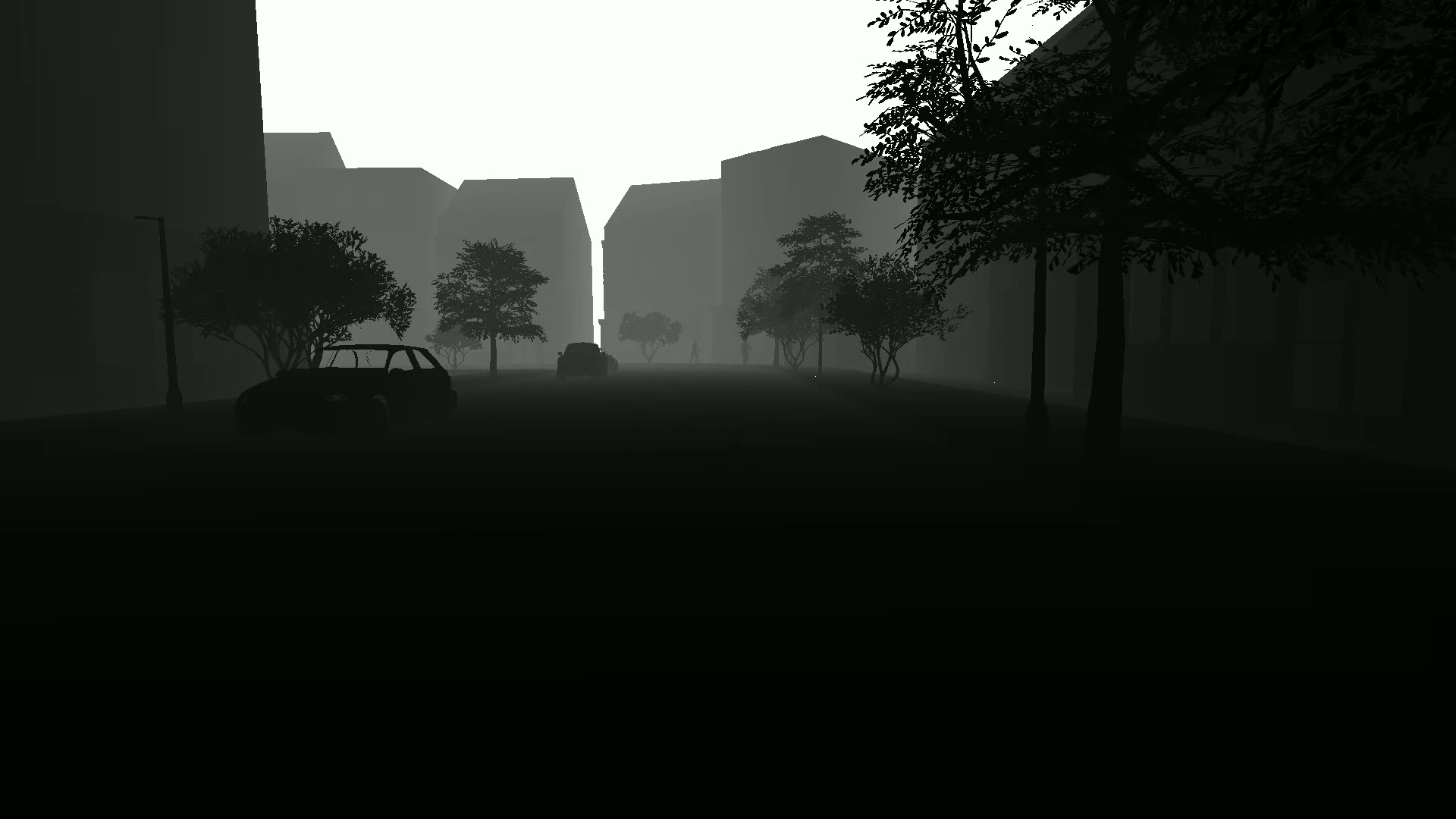}
			\includegraphics[width=0.23\textwidth]{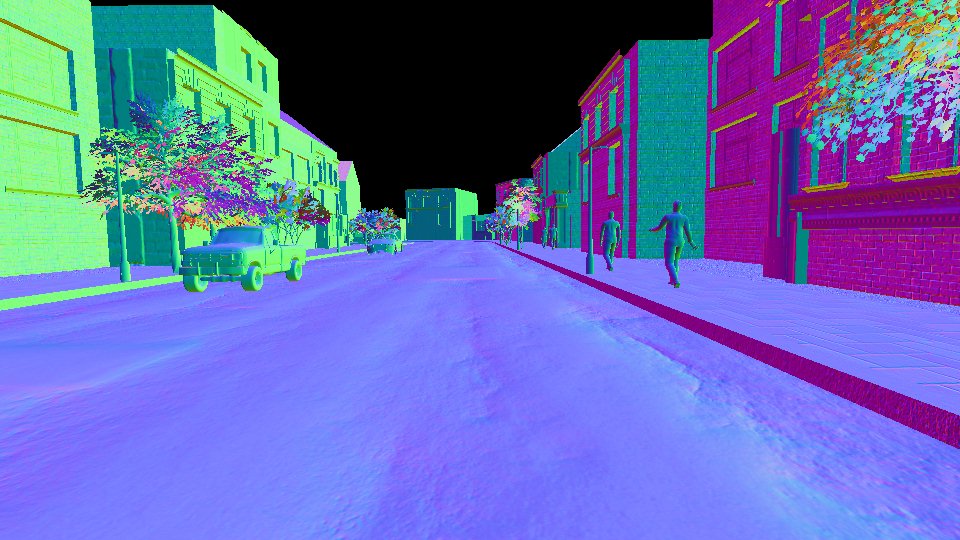}
		\end{subfigure}
		\caption{Example video stream snapshots. 
			Row A illustrates common environmental changes, such as variations in illumination and weather conditions. Row B depicts two potential examples for class incremental learning, where entire object classes, here trees, progressively appear or disappear (left image pair), or the learning task is based on successive availability and granularity of  annotations (right image pair).
			Row C shows exemplary de-activation of specific physics-based material properties such as color, surface normals, or (sub-)surface reflections, resulting in a fully gray or colored flat shaded world, without any reflection or small cavity details (left image pair). Built-in availability of additional depth and surface normal annotations is further highlighted (right image pair).
			\label{figure:render_examples}}
	\end{figure}
	
	Inspired by previous works in the context of urban scene segmentation \cite{Geiger2013,Ros2016,Cordts2016,Richter2016}, our primary contribution is the introduction of a modular Unreal Engine 4 \cite{EpicGames2015} based 3-D computer graphics simulator that now also enables clear-cut generation and assessment of diverse continual learning scenarios. A selection of video snapshots is illustrated in figure \ref{figure:render_examples} and additionally a showcase video can be viewed at \url{https://youtu.be/8zDhol8CIf0}. Specifically, we:
	\begin{itemize}[leftmargin=0.5cm,labelwidth=\parindent,labelsep=2pt]
		\item Introduce a simulator that facilitates grounded investigation of continual learning mechanisms through access to highly customizable data. At all times, our simulator only renders an upcoming segment of the world through efficient real-time scene assembly. Its offered data generation is based on manipulation of temporal priors and parameters of the generative model. Our modular control spans aspects from physics-based (de-)activation of color, surface normals and scattering, to switches in weather conditions or environment lighting, and ultimately to commonly evaluated abrupt changes in the data population though (dis-)appearance of entire object categories. 
		\item Corroborate our simulator's utility in an initial showcase of multiple continual learning techniques, investigated across incremental class, environmental lighting, and weather scenarios. 
		\item Provide open access to:  \\
		1. As a benchmark creation tool, a stand-alone simulator executable with configuration files for the specification of rendered sequences: \url{https://doi.org/10.5281/zenodo.4899294} \\
		2.  To allow extensions, the underlying source-code of the simulator:\\ \url{https://github.com/ccc-frankfurt/EndlessCL-Simulator-Source} \\
		3. A set of respective videos and their precise dataset versions to reproduce the particular experiments of this paper: \url{https://doi.org/10.5281/zenodo.4899267}.
	\end{itemize}
	We have made use of the Zenodo platform to ensure persistence of our datasets and software, while also making sure that our content has a DOI with versioning capabilities for  future updates.
	
	\section{Endless Procedural Driving Simulator}
	Our procedural world generation framework allows for creation of temporally consistent video streams, where respective sub-sequences are subject to an interpretable parametric generative model through which the scenario is continuously adaptable. This can be gradual and seamless changes in the environment, or mirror abrupt shifts in the world's configuration. Our specific implementation is inspired by urban driving. Our considered main actor is a \textit{vehicle} with statically attached \textit{camera} that drives along a procedurally generated track of successive \textit{street segments}. Every such street segment is randomly selected to balance curvature and crossings with straight roads. For each sampled street element, various \textit{objects}, such as \textit{buildings}, \textit{trees} or \textit{street lamps} are stochastically placed according to real-world motivated priors. Additional \textit{dynamic actors}, i.e. \textit{other vehicles} and \textit{humans} are sampled for each segment, their \textit{motion} dynamics and \textit{animations} drawn at random. At all times, the number of existing street segments remains constant. As the main camera actor proceeds through the world, novel segments are procedurally generated, while already observed ones are disposed of. Through control of spawning probabilities or parameters administering \textit{weather} and \textit{lighting}, the user is granted the ability to adapt the upcoming world of the real-time generated video.
	
	\subsection{Generative Model}
\begin{figure}[t]
		\centering
		\resizebox{0.975 \textwidth}{!}{%
			\includegraphics[width=1\textwidth]{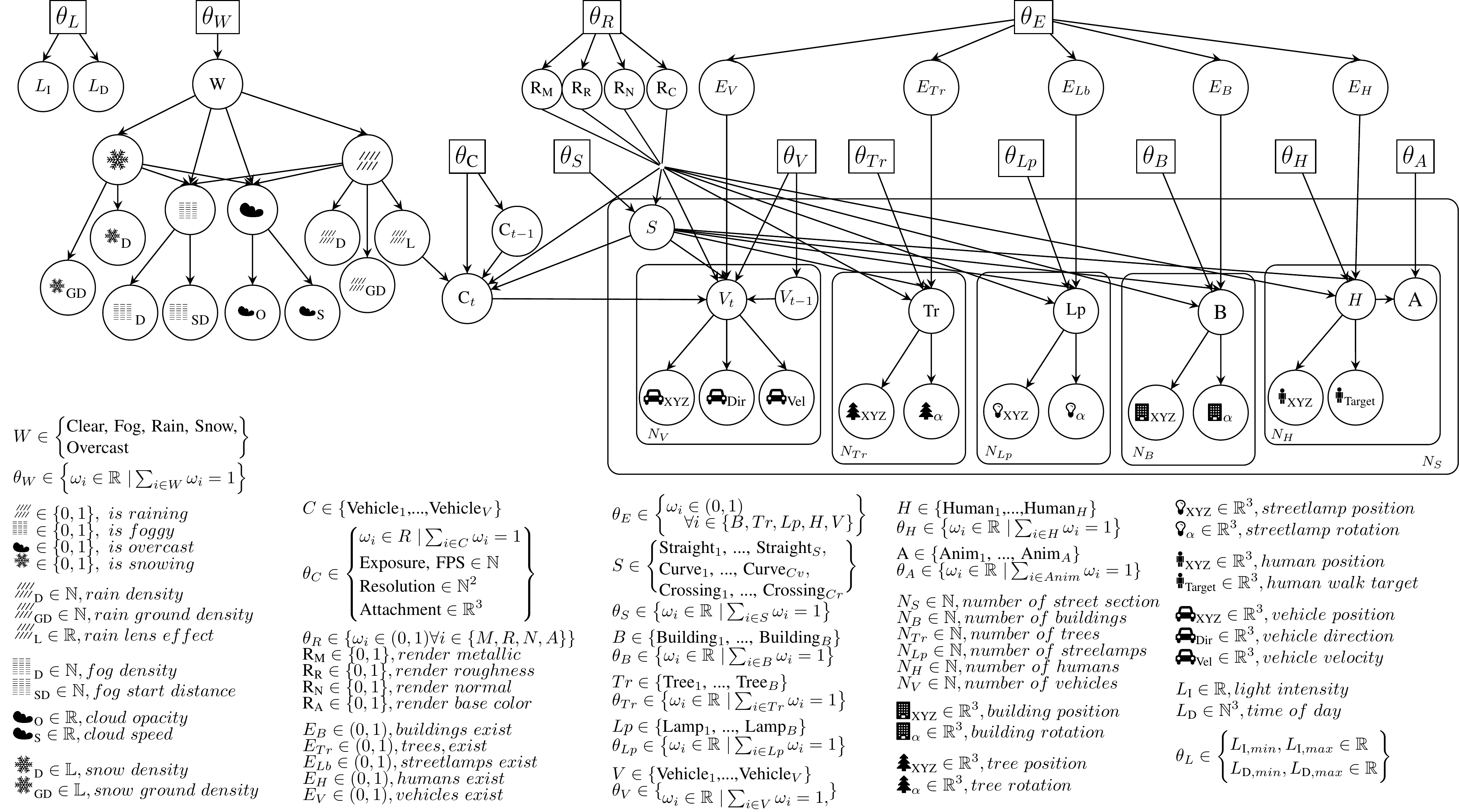}
		}
		\caption{Parameterized probabilistic graphical model. Plate notation \cite{Buntine1994} is used for repeated random variables, sets of parameters $\mathbf{\Theta}$ are denoted by rectangles. As the generative process creates entire video sub-sequences, only the camera $C_{t}$ and vehicle actors $V_{t}$ have an explicit dependency between two consecutive sampling steps. All other random variables are constrained to reside in their respective sub-sequence and consecutively sampled independently. The subscript $t$ has thus been omitted for ease of readability. The physics-based rendering random variables $R_{M,R,N,C}$ are each connected to all objects and actors. Their arrows have been merged visually to avoid excessive clutter.\label{figure:generative_model}}
	\end{figure}

	We define an entire video sub-sequence as a random variable $X_t = \left\langle \text{S,B,Tr,Lp,H,V,C,W,L,E,R} \right\rangle$, with street segment S, static buildings B, trees Tr, and street lamps Lp, dynamic human and vehicle actors H and V, a car with attached camera C, weather condition W and lighting L. A meta-variable E governs the existence of entire object and actor categories and R controls the physics-based material rendering model. A continuously growing video is therefore comprised of $t = \{1,...,T\}$ sub-sequences,  defined by the parametrized probabilistic generative process given by:
	\begin{align}
		P(X_t|\Theta) = &P(V_t | V_{t-1}, S_t, \theta_{V, t}, E_{V, t}, R_t) \ P_t (B, Lp, Tr | S, \theta_B, \theta_{Tr}, \theta_{Lp},  E_{B}, E_{Tr}, E_{Lp}, R)\nonumber\\
		& P(C_t | C_{t-1}, \theta_{C, t}, S_t, R_t) \ P_t(A | \theta_A, H) \, P_t(H | S, \theta_H, E_{H}, R) \\
		& P_t(E_{B},E_{Tr},E_{Lp},E_{H},E_{V} | \theta_E) \ P_t(S | \theta_S, R) P_t(R | \theta_R) P_t(W | \theta_W)  \ P_t(L_\text{I}, L_\text{D} | \theta_L) \nonumber .
	\end{align}
	Here, notation is simplified for $R = \{R_{M}, R_{C}, R_{N}, R_{R}\}$, manipulating a material's metallicity, color, surface normals and roughness, and $P_{t}$ is short for an index $t$ at every entry of an expression. All variables and their stochastic dependencies are illustrated in the graphical model of figure \ref{figure:generative_model}.
	We proceed by elaborating on the random variables and their distributions parametrized by respective $\boldsymbol{\Theta}$.
	
	\subsection{Random Variables and Distributions}
	The random variables of our generative process are subject to three families of distributions: discrete categorical or Bernoulli, and continuous distributions on bounded intervals. The former two generally reflect a random selection of e.g. object or actor styles, and their general existence and co-occurrence. The latter expresses our belief in e.g. plausible object or actor locations through parametrization of the finite support and their overall amount. With few exceptions and as will be detailed in an instant, these variables are presently assumed to originate from uniform distributions, but can generally be sampled with more complex distributions and interdependencies, if for instance a specific city composition is desired to be emulated. 
	
	\textbf{Street segments, main actor and vehicles:} The overarching categorical random variable $S$ indicates the stochastic selection of street-segment layouts, i.e. elements containing various straight and curved road designs, or crossings, with respective sampling probability values given by $\theta_S$. The segments themselves are not placed stochastically. Instead, the beginnings of consecutive sequence elements are deterministically attached to their predecessor's end, in order to seamlessly continue the endless procedural generation of the world. The main actor of a vehicle with statically coupled camera, random variable $C$, follows this world as if it were on a pre-determined track, i.e. the direction of the camera vehicle is thus contained in the stochastic sampling of street segments. $C$ itself is of categorical nature and represents a choice in vehicle model, with additional $\theta_C$ describing camera properties such as resolution, exposure or the captured video's frames per second. In principle, it is not directly subject to a random velocity. However, the velocity does vary in dependency on other stochastic variables. These include further dynamic vehicles, expressed through the categorical variable $V$ and its parameters $\theta_V$. Again, these vehicles are randomly chosen from the repertoire of vehicle models. Their location is sampled from a uniform distribution on a bounded interval that is constrained to the extent of the sampled street segment, their velocity selected at uniform in the range of assumed minimally and maximally realistic velocities in an urban scenario. If these vehicles drive slowly in front of the camera main actor vehicle, they inevitably lead to deceleration in order to avoid collision. As occasional auxiliary dynamic vehicles can thus continue their path and persist across multiple video sub-sequences, they represent the only random variable apart from the main camera that relies on its former sub-sequence's state. 
	
	\textbf{Static objects and human actors:} All other dynamic and static random variables are contained within their respective street segments. These include a population of possible static 3D-objects, presently captured by the set $Obj = \{B, Tr, Lp\}$, corresponding to buildings, trees and street lamps, and additional dynamic human actors $H$. Again, the respective categorical random variable indicates a choice in available models. Human actors are further dependent on categorical random variable $A$, enabling the random choice from a discrete set of animation. Once more, their sampling probability is given through respective parameters, and their stochastic initial and final pose, representing the walk target, is presently drawn from uniform distributions with finite support, determined by the scope of the street segment on which they are randomly placed. 
	
	\textbf{Category existence:} With the exception of the always present world's street segments and tracked camera vehicle, all previously introduced categorical random variables are modulated through Bernoulli variables on a top-level. The corresponding $E_{i} \, \forall i \in \{B, Tr, Lp, H, V \}$, parametrized by $\theta_E$, govern the overall "existence". In a sense this is a convenience random variable, that allows trivial control over the presence of entire categories, in foresight of class incremental scenarios where e.g. all trees appear or disappear when driving through or leaving an avenue. In general, this could have been adjusted through the number of spawned objects and actors, i.e. the repeated sampling of random variables for one sub-sequence expressed through plate notation \cite{Buntine1994} in figure \ref{figure:generative_model}. For ease of readability, we define the respective random variable $N_i \, \forall i \in \{B, Tr, Lp, H, V\}$ independently and for now assume that distinct amounts, limited by an empirical $N_{i, max}$, are a priori equally likely.
	
	\textbf{Weather and lighting:} A complementary part of the graphical model controls weather and lighting. Categorical $W$, parametrized by $\theta_W$, represents an initially mutually exclusive choice between five distinct weather conditions: clear, fog, rain, snow and overcast (cloudy). Depending on the concrete outcome, an additional set of Bernoulli variables covers potential co-occurrence of fog and clouds when rain or snow are active. Each of these conditions is subject to further random variables on density, ground density (for snow cover and puddles), and camera lens effects. For now, these latter variables are sampled from uniform distributions ranging from zero to unity in terms of effect strength. In complete analogy, lighting for a video sub-sequence is given by continuous random variables on intensity and daytime, defining the illumination angle and color due to atmospheric scattering. 
	
	\textbf{Physically-based rendering:} Finally, the world is assumed to follow a physics-based rendering process. Parametrized by $\theta_R$, a set of Bernoulli random variables $R_r \, \forall r \in \{M,R,N,C\}$ dictates the activation of rendering aspects with respect to all materials, i.e. metallicity, surface roughness, normals, and base color. These define how material appearance manifests in the observed video-subsequence, with potential to exclude color or particular surface or subsurface light interactions. 
	
	\subsection{Practical Details}
	\begin{figure*}
		\centering
		\begin{tabular}{c c c}
			\includegraphics[height=0.15\textheight]{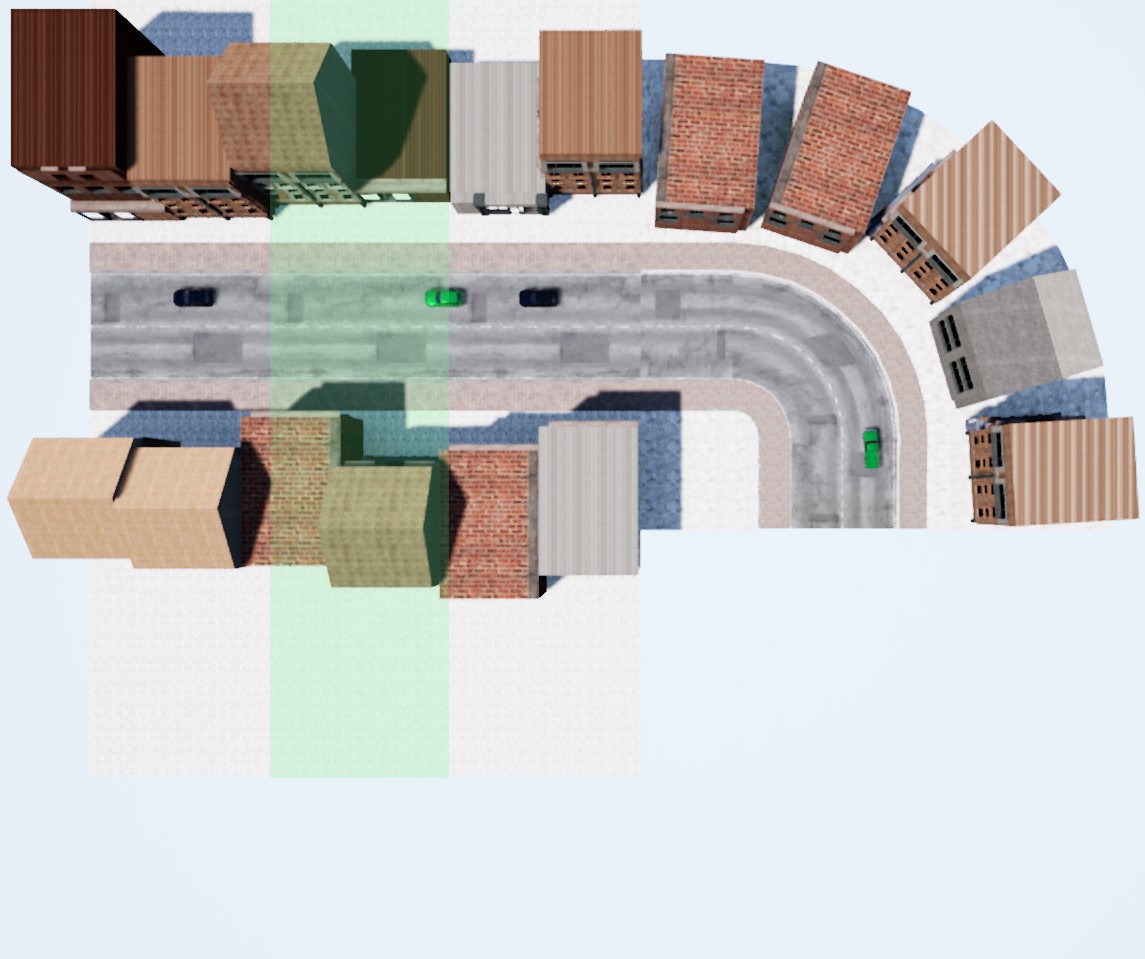} & 
			\includegraphics[height=0.15\textheight]{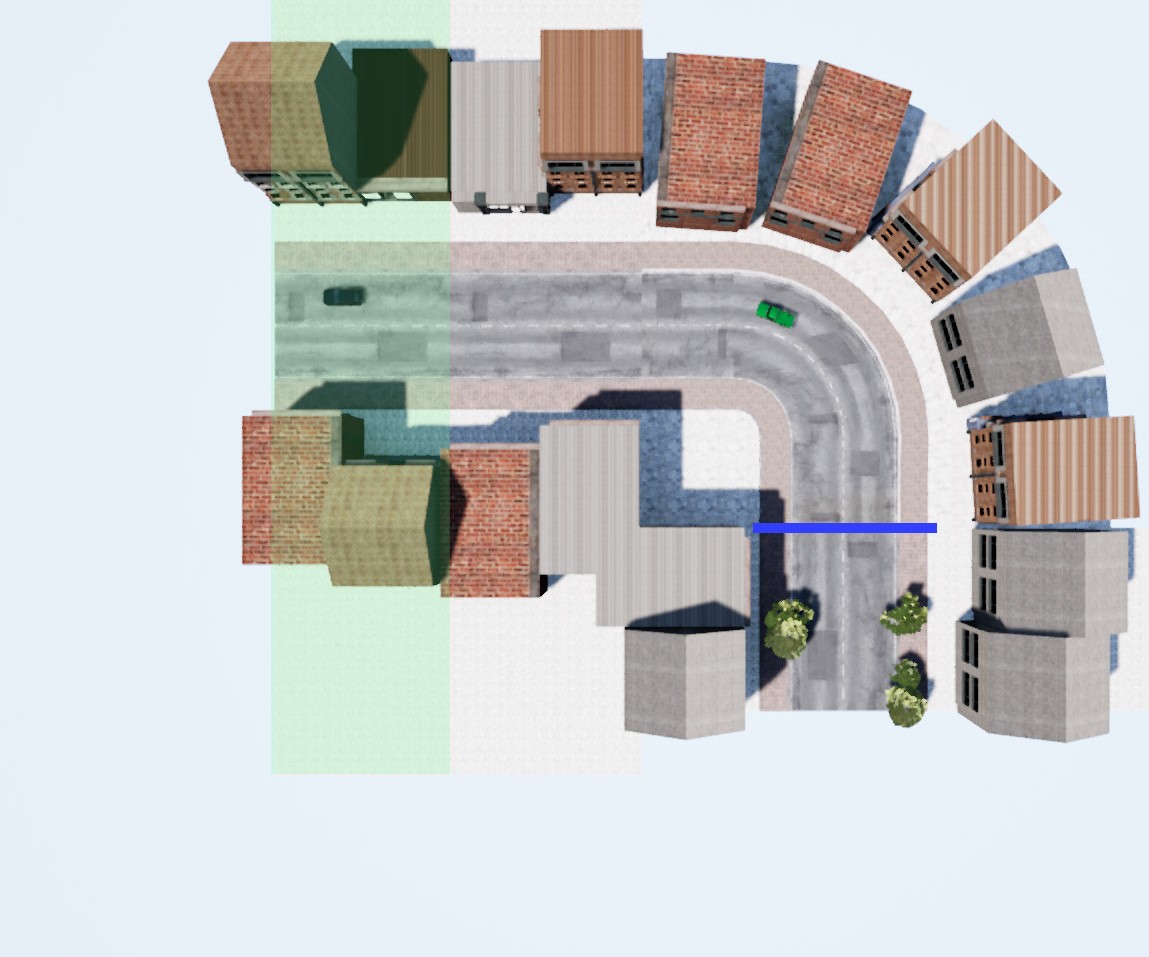} &
			\includegraphics[height=0.15\textheight]{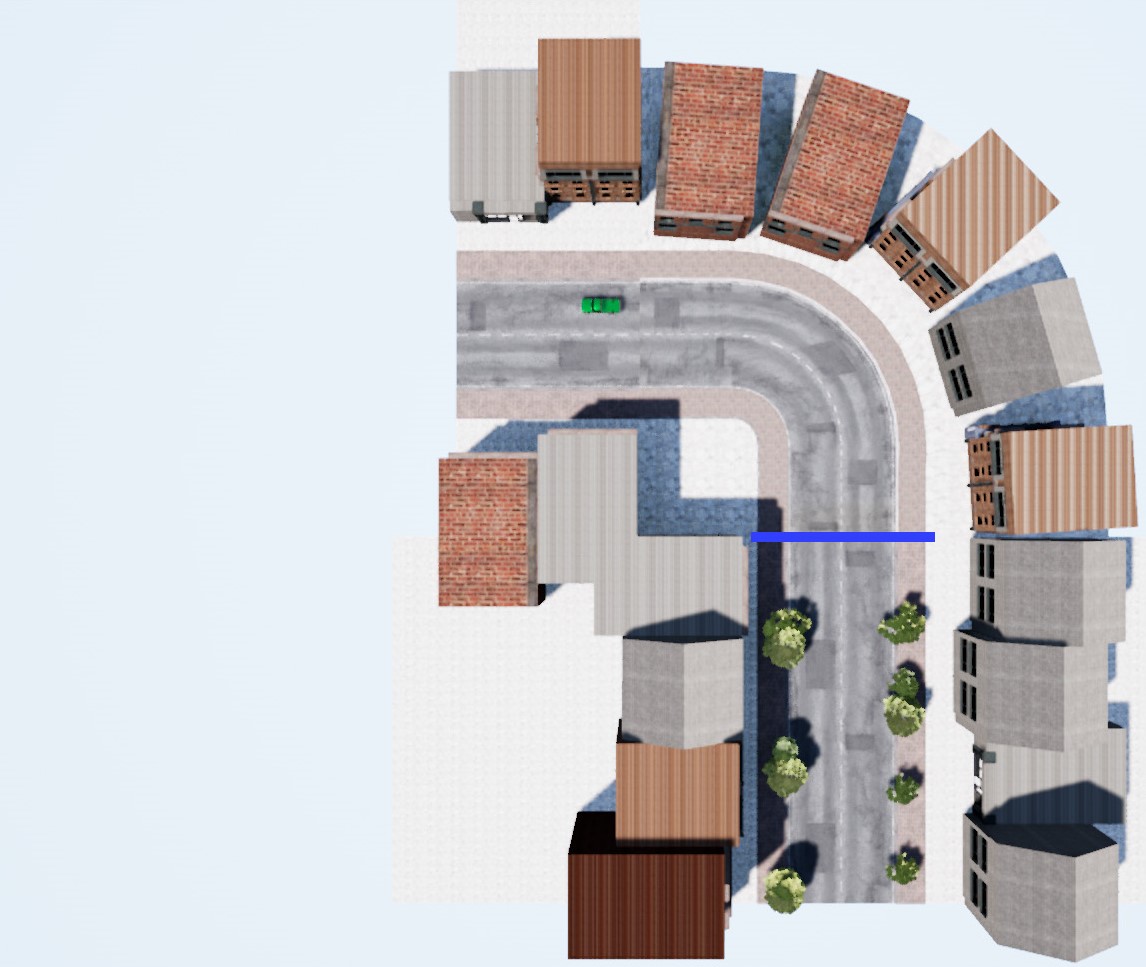}\\
			Sub-Sequence 1, Tile 4 & Sub-Sequence 2. Tile 1 & Sub-Sequence 2, Tile 2
		\end{tabular}
		\caption{Birds eye view of the stochastic procedural scene generation. Switching the scene's configuration from 'Sub-Sequence 1` to 'Sub-Sequence 2` it can be observed how the object population changes to exclude cars and feature trees in the newly added tiles. The area highlighted in green marks the extent of a single tile and the blue line indicates the seam between the two sub-sequences.\label{figure:tiling}}
	\end{figure*}
	The practical simulation is based on Unreal Engine 4 \cite{EpicGames2015}, supporting our requirements for specification of distributions through code and dynamic world generation, while aiming for real-time high photo-realism. The essential street segments are represented through a discrete set of tiles, where for each tile type spawn volumes govern the finite support bounds of the underlying location sampling for each object and actor. In addition, each such tile includes anchor points for attachment with further tiles. 
	Figure \ref{figure:tiling} shows a birds eye perspective of a constructed world segment and illustrates the procedural scene generation process. For details on assets, distribution parameter choices, and the spawning bounds we refer to the appendix.
	The use of tiles elegantly encapsulates the generative model's probabilistic sampling properties in practice, while yielding considerable memory and computation benefit by strictly limiting the amount of resources to be managed at each time step. 
	
	We simultaneously capture color video, semantic pixel annotations, depth, and normals. Continuous world generation with capture of a video with half-HD ($960\times540$) resolution, from a single camera, achieves $\sim 15$ frames per second (fps) on a consumer GTX-1080 GPU. With only the original video and one additional mode, this number increases to $28$ fps. In either way, we allow the user to set a desired capturing rate and make use of time dilation, i.e. virtually slowing down the scene, if the same world segment is needed to be captured with higher frame rate or extensive rendering features such as above full-HD resolutions. 
	As pointed out in our contributions, we provide a stand-alone simulator executable. Here, used content is encrypted and can be used without explicitly sharing licensed assets. Arbitrary chains of distribution modifications for video sequences to be generated can be specified through a JSON-config file. For inclusion of novel assets we point the reader to our shared source-code. We provide more detailed descriptions and usage instructions in the supplementary.
	
	\section{Related Work}
	The idea to leverage synthetic data creation for the training of deep neural networks has seen an impressive amount of successful practical implementations \cite{Li2012,Satkin2012,Vazquez2014,Peng2015,Hattori2015,Handa2016,Richter2016,Dosovitskiy2017,DeSouza2017,Jiang2017,Saleh2018}.
	The common expectation is that learned appearance models can be adapted in their domain to an ultimately desired real world target task. Various works have therefore focused on automated calibration based on similarity computations between virtual and real images \cite{Satkin2012, Vazquez2014}, manifold alignment to match scale between generated and synthetic texture data \cite{Li2012}, or adversarial tuning to assimilate data population statistics \cite{Veeravasarapu2017, Shrivastava2017}. On the alternative end of the respectively constructed scenes, the general aim is thus to augment real world datasets \cite{Peng2015}, customize aspects that are otherwise difficult to capture, such as pedestrian motions \cite{Hattori2015} or human actions \cite{DeSouza2017}, in an effort to overcome laborious human annotation in creation of large-scale datasets with massive amounts of variation. Depending on specific works, the spotlight can be on e.g. fixed urban scenes and variations of environmental factors \cite{Ros2016, Dosovitskiy2017}, direct extraction of such scenes from video games \cite{Richter2016}, or the complete randomization of all factors to maximize the amount of conceivable configurations \cite{Handa2016, Jiang2017, Veeravasarapu2017}. Whether or not all factors are randomized in stochastic processes or scene elements remain static, the generally measured utility and impact is derived from measurements on corresponding real world benchmark datasets \cite{Cordts2016, Xiao2010, Cordts2016, Russakovsky2015, Geiger2013}.
	
	Although valuable for their proposed purpose, we posit that existing simulators are not natural for exploratory study of continual learning limits. In simulators that use pre-determined layouts the user is limited to specific geometrical configuration and scene types. If assumed static buildings and other actors need to be replaced, removed or complete object categories are desired to be added, the world needs to essentially be recomposed \cite{Hattori2015, Richter2016, Ros2016, Dosovitskiy2017}. Analogously, works that employ stochastic point processes or similar hierarchical procedures to randomize entire scene configurations, including stochastic camera placement to vary frames or locally consistent video segments \cite{Ros2016, Veeravasarapu2017, Hess2017, Jiang2017, DeSouza2017}, require significant amounts of compute in recurring composition of the entire world if the user wishes to change priors on underlying generative factors. As an additional challenge this is only possible if, and only if, the associated simulator core and source code has been publicly released beyond an executable, because the desired settings are typically not applicable through the provided user interface. Our proposed simulator differentiates in this regard, as we can dynamically change parameters in real-time generation to e.g. spawn trees in the distance, de-spawn specific buildings, change locations, vary lighting and modulate weather effects. Such nuanced adaptability of all scene elements enables easy creation of a dynamically adaptable endless procedural world.
	
	From the perspective of continual learning, our described simulator thus allows for composition of data sequences in an effort to extent presently limited analysis. In particular, we can enable analysis beyond present continual vision practice, that rests primarily on not well understood sequentialization of popular classification benchmarks \cite{LeCun1998,Nilsback2006,Krizhevsky2009,Xiao2010,Netzer2011,Russakovsky2015}. The latter typically undergo class specific splits, permutation, concatenation or other alterations through augmentation to provide pre-designed iterative sequences of object or class information \cite{Lomonaco2017,Rebuffi2017a}. Based on these contrived and uncontrolled benchmarks multiple families of continual learning approaches have formed. These range from simple rehearsal of original data subsets \cite{Ratcliff1990,Rebuffi2017,Isele2018,Rolnick2018} to generative data replay \cite{Robins1995,Shin2017,Mundt2020a}, or from functional regularization \cite{Li2016,Rannen2017,Zhai2019}, based on knowledge distillation \cite{Hinton2014}, to explicitly constraining parameters \cite{Zenke2017, Kirkpatrick2017}. Unfortunately, in empirical comparison, it quickly becomes apparent that assumptions of individual methods are narrow and seem to often be practically tailored towards the limited use case of a particular benchmark \cite{Pfulb2019,DeLange2019,Lesort2019, Diaz-Rodriguez2018, Kemker2018, Lopez-Paz2017, Farquhar2018a, Mundt2020b}. We would argue that this is not necessarily a direct result of originally misguided design, but rather a consequence of the underlying original datasets being seldomly designed with continual learning in mind. Our works imminent goal is thus to deepen our understanding of when and why deep learning fails in continuous training, how potential curricula impact learning, and how mechanisms can be improved to consistently mitigate shortcomings across a wider range of scenarios.

	\section{Deep Continual Learning Experiments} \label{section:experiments}
	We empirically corroborate our simulator's utility in a set of initial experiments. Here, we showcase that catastrophic interference can still be a significant challenge across many literature methods, even when only considering simulated data. Inspired by the typically limited evaluation of deep continual learning in class incremental scenarios \cite{Pfulb2019,DeLange2019,Lesort2019,Diaz-Rodriguez2018,Kemker2018, Farquhar2018a, Lopez-Paz2017,Mundt2020b}, we now generate and investigate video sequences in three distinct set-ups: \textit{incremental class appearance}, \textit{varying weather conditions} and \textit{decreasing illumination intensity}. They have been selected to display the benefits and shortcomings of currently prevalent techniques to alleviate catastrophic interference, and consequently why it is necessary to make use of our simulator for a more diverse evaluation. 
	
	For this purpose we consider popular approaches from various families of continual learning mechanisms: synaptic intelligence (SI) \cite{Zenke2017} and elastic weight consolidation (EWC) \cite{Kirkpatrick2017} for parameter regularization, functional regularization through knowledge distillation as presented in learning without forgeting (LwF) \cite{Li2016}, as well as data replay methods. For the latter, we consider replay using gradient episodic memory (GEM) \cite{Lopez-Paz2017}, a straightforward exemplar rehearsal mechanism, where a subset of data is stored and interleaved in continuous training in the spirit of \cite{Nguyen2018, Rebuffi2017}, and generative pseudo-replay with open set classifying denoising variational auto-encoder (OCDVAE) \cite{Mundt2020a}.
	
	\subsection{An Initial Set of Considered Scenarios}
	For our investigation, we have selected the simplest conceivable task of classification, where all objects' bounding boxes are assumed to be detected perfectly. Even in this significantly facilitated setting we will see that many investigated techniques are more brittle than desired. We emphasize that our simulator is naturally capable of rendering data for more complex object detection, surface normal or semantic segmentation investigations. The detailed generated video sequences are categorized according to three scenarios:
	
	\textbf{Incremental Classes:} Representing the most commonly investigated continual scenario, our video-stream consists of four video sub-sequences, each adding one distinct object class to the task. In the training set, each sub-sequence contains only one object category, where buildings $B$ in conjunction with the street-section itself are attributed to an always present 'background' class. In contrast, a separately generated test set progressively accumulates all present classes. Recall, that we can express this change throughout the video sequences with $\pi_{E, t}$. The temporal sequence of chosen Bernoulli likelihoods for the vector of buildings, trees, street-lamps, humans and vehicles is $\pi_{E, t=1} = (1,1,0,0,0)$, $\pi_{E, t=2}  = (1,0,1,0,0)$, $\pi_{E, t=3}  = (1,0,0,1,0)$, $\pi_{E, t=4} = (1,0,0,0,1)$. We note that after initially choosing weather and lighting conditions for $t=0$, we further adapt their parameters to have consistent weather and lighting for the remainder of sub-sequences.
	
	\textbf{Incremental Lighting:} The incremental lighting video sequence is based on progressive decrease in illumination intensity as a single generative factor, without adjustments to the illumination color. The generated training video stream consists of five sub-sequences, where the typically sampled uniform distribution for light intensity is collapsed to a sequence of delta distributions with concentrated mass at  $\pi_{L, t=1} = 76.8$, $\pi_{L, t=2} = 19.2$, $\pi_{L, t=3} = 9.6$, $\pi_{L, t=4} = 2.4$, $\pi_{L, t=5} = 1.2$, expressed in units of Lux. The test simply consists of a growing amount of separately generated sub-sequences through time. The categorical weather distribution is parametrized to have a probability of one for clear day and all object categories exist at all times, with all locations and total amounts sampled stochastically.
	
	\textbf{Incremental Weather:} Incremental weather video streams can be defined in terms of probabilities on the categorical weather variable. At each point in time, we set the probability for a specific outcome to 1 and all other choices to 0. The corresponding temporal sequence for occurrence of clear, rain, snow, fog, overcast conditions is then 
	$\pi_{W, t=1}= (1,0,0,0,0)$, $\pi_{W, t=2} = (0,1,0,0,0)$, $\pi_{W, t=3} = (0,0,1,0,0)$, $\pi_{W, t=4} = (0,0,0,1,0)$, $\pi_{W, t=5} = (0,0,0,0,1)$. Probabilities for the Bernoulli variables defining the existence of object types are all set to unity, such that all objects get sampled at all times. The test video stream is defined in analogy to the incremental lighting scenario.
	
	\subsection{Experimental Setup and Evaluation}
	Each scenario is captured in a video sequence of $960 \times 540$ resolution, consisting of multiple approximately 15 minutes long sub-sequences with 150 sampled street segments, and a respective test set video.
	We base encoders, and the VAE decoder, on the popular four convolutional layer architecture of Radford et al. \cite{Radford2016}, without temporal dependency and thus a frame-wise prediction. In all our models we make use of a single classification head for all tasks in the presented sequence. 
	
	We presently focus on monitoring of simple classification accuracy and train the neural networks to full convergence on a sub-sequence before proceeding. 
	That is, the train sets only ever consist of data from the current sub-sequence/time-step. In contrast, the test set accumulates data successively. To give an example based on the  class incremental scenario, the train set of task 1 thus consists of the omnipresent background class and trees, whereas task 2 consists of data featuring background and cars. In contrast, the test set accumulates observed classes and the test accuracy is measured over all classes seen up to the present time step, i.e. task 2 would classify background, trees and cars. This procedure mimics the typically conducted evaluation in class incremental learning scenarios, in which the test performance provides a rather direct indication whether the former tasks are being catastrophically forgotten in continuous optimization.
	
	As such, the overall objective grows in complexity over time, along with the number of tasks presented. The accuracy of the investigated continual learning techniques is thus compared to the maximally achievable upper-bound accuracy, assuming the upfront presence of the entire video, and a naive continued training, where training greedily continues only on the current sub-sequence without any mechanism to prevent catastrophic interference. 
	To provide information on the statistical deviation of all approaches, each experiment has been repeated $5$ times. A detailed account of the training, it's hyper-parameters and how datasets were generated can be found in the appendix. Our code is an extension of the public OCDVAE codebase \cite{Mundt2020a} in combination with the Avalanche \cite{Lomonaco2021} continual learning library. It is available at: \url{https://github.com/TimmHess/OCDVAEContinualLearning}. 
	
	\subsection{Continual Learning Results}
	\begin{figure*}[h]
		\centering
		\hspace*{-1.5em}
		\begin{tabular}{p{0.305\textwidth} p{0.305\textwidth} p{0.305\textwidth}} 
			\includegraphics[width=0.33\textwidth]{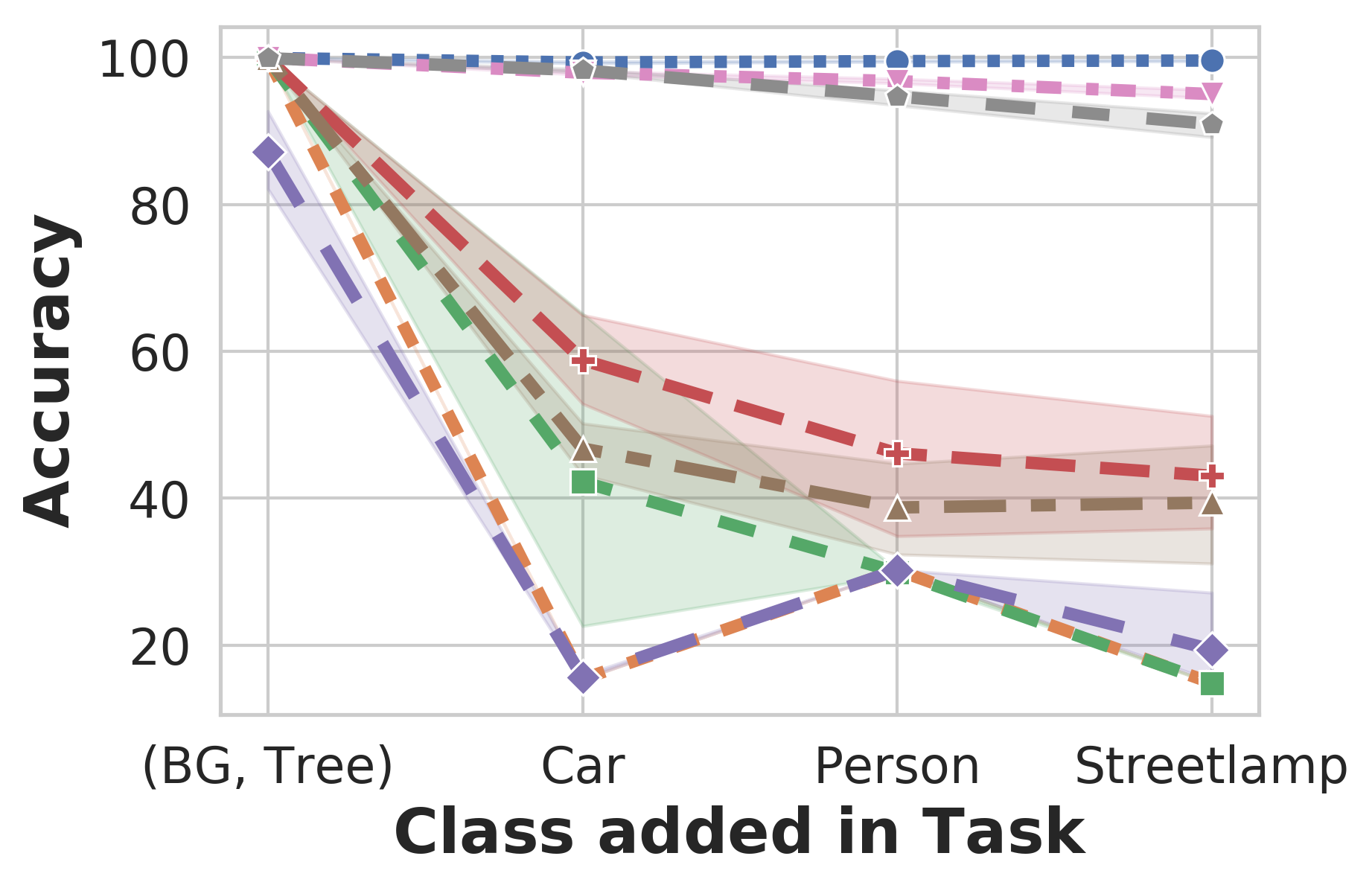} &
			\includegraphics[width=0.315\textwidth]{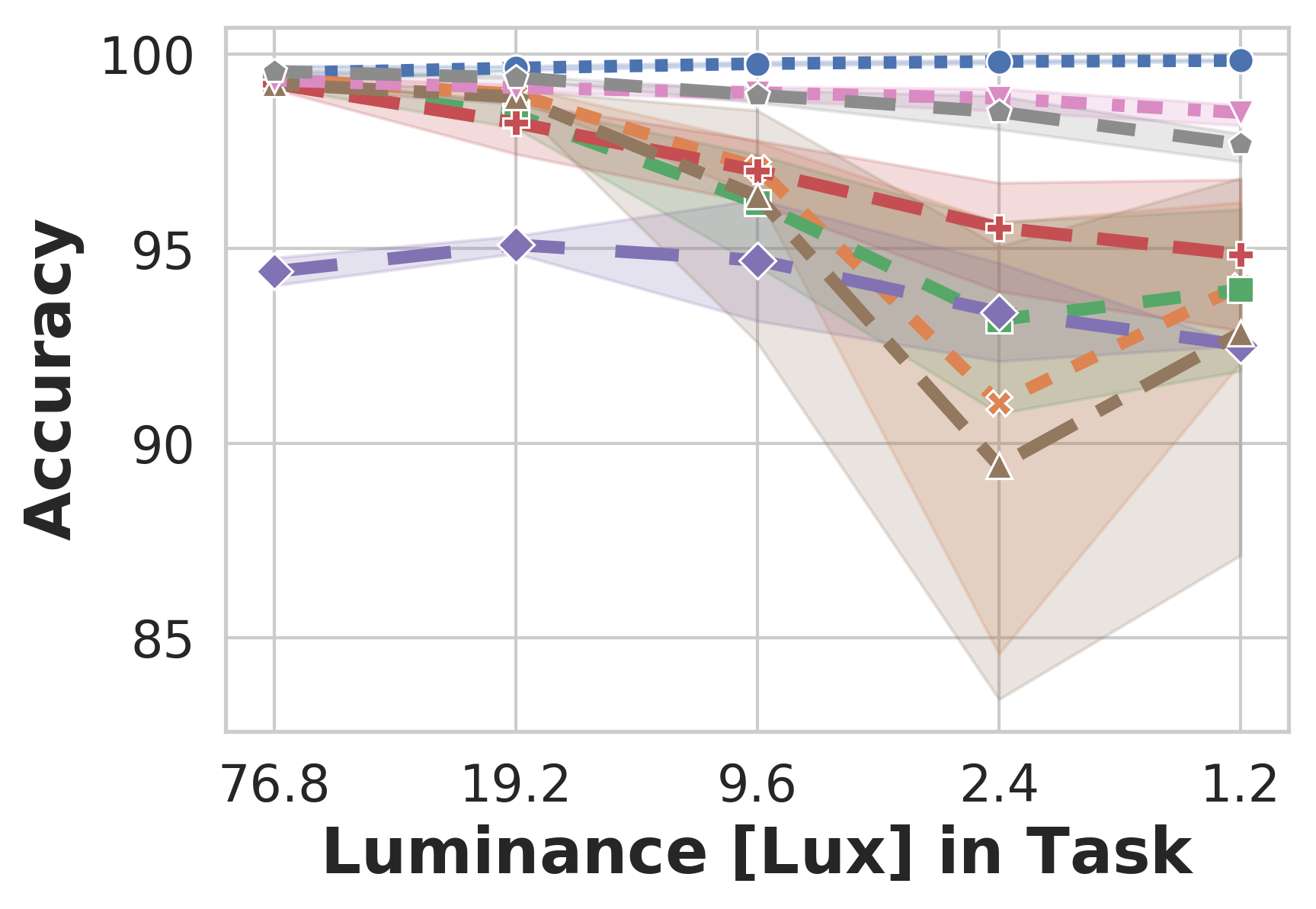} &
			\includegraphics[width=0.33\textwidth]{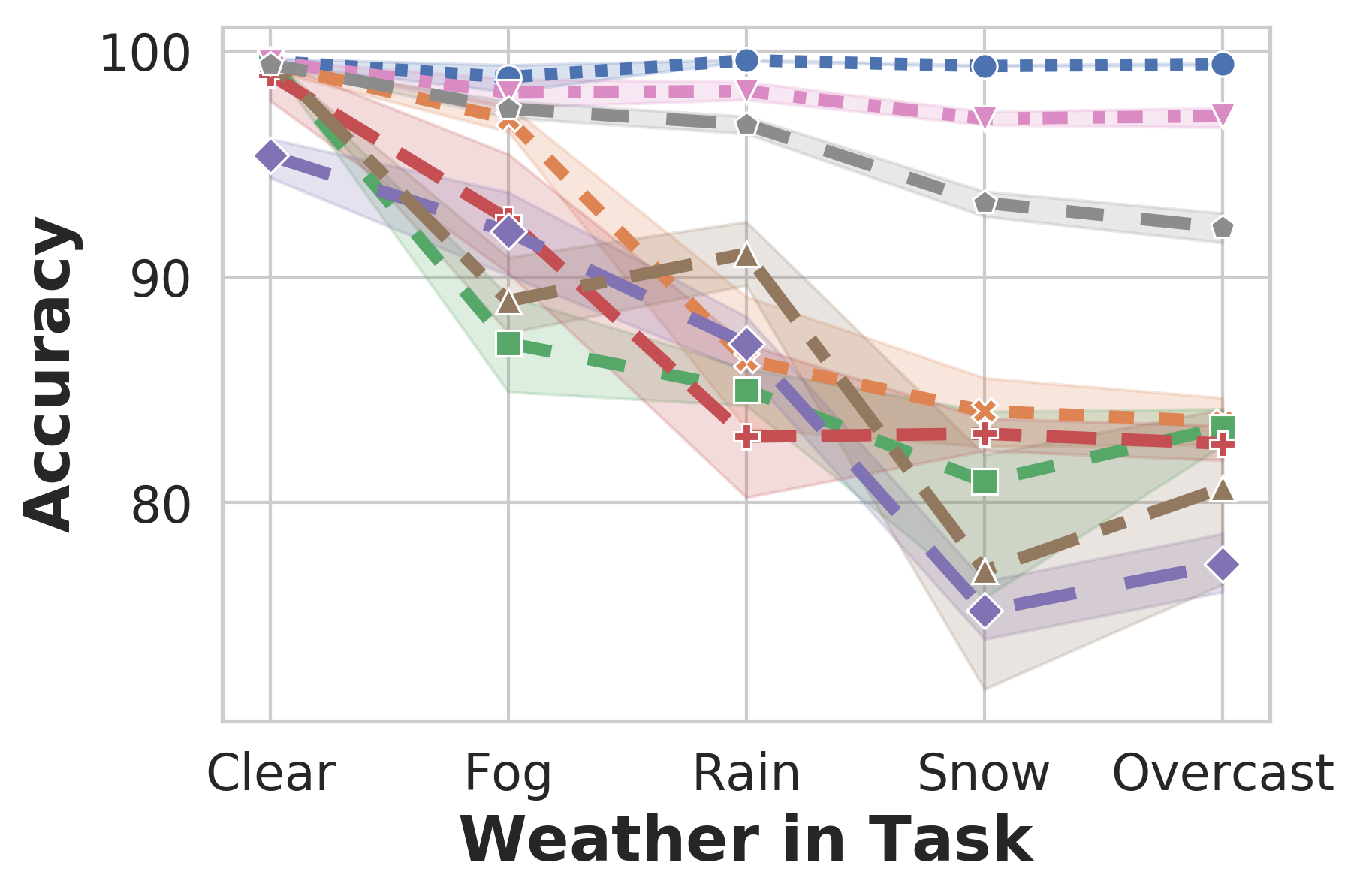}\\
			\multicolumn{3}{c}{
				\includegraphics[width=0.8\textwidth]{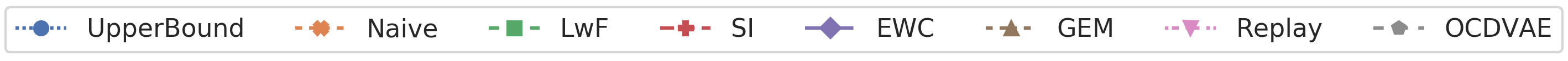}
			} 
		\end{tabular}
		\vspace*{-0.5em}
		\caption{Comparison of deep continual learning accuracy across five experiments in three conceivable scenarios: incremental object appearance (left), decreasing illumination intensity (middle) and changing weather conditions (right). 
			Accuracy for each method is measured after completion of the training phase for every respective task increment. The training data stems only from the current task increment to be learned, while evaluation is conducted on the cumulative test data comprising all tasks up to the present point in time.
			Learning without Forgetting (LwF) \cite{Li2016}, Synaptic Intelligence (SI) \cite{Zenke2017}, Elastic Weight Consolidation (EWC) \cite{Kirkpatrick2017}, Gradient Episodic Memory (GEM) \cite{Lopez-Paz2017}, direct exemplar replay (Replay), and Open-set Denoising Variational Auto-Encoder (OCDVAE) \cite{Mundt2020a} are contrasted with naively continued training (Naive) and the maximally obtainable accuracy when accumulating all tasks' data (UpperBound).
			\label{figure:CL_results}}
	\end{figure*}
	Figure \ref{figure:CL_results} shows the overall achieved accuracies at the end of each sub-sequence for the considered continual learning techniques, corresponding naive continued training, and the maximally attainable upper-bound. Note that the upper-bound indicates that the three scenarios can in principle be fully solved with the chosen neural architecture. Whereas regularization methods provide some benefit in class incremental scenarios, they seem to fall behind even a naive continued training, which is supposed to incur full catastrophic interference, in the other settings. For instance, they all significantly under-perform with global weather changes. Although the employed OCDVAE generative replay can be observed to prevent the occurrence of catastrophic interference almost completely for class increments, it also begins to struggle when the environment's weather changes. We suspect that this might be due to the generative model having more difficulties in capturing the statistics of weather and its implications such as puddles, in contrast to only having to accurately generate different objects under fixed environmental conditions. 
	
	Interestingly, the trends observed in regularization techniques seem to be mirrored when using gradient episodic memory, even though it makes use of auxiliary stored data examples. We hypothesize that this is a consequence of GEM nevertheless relying on a regularization technique at its core, that is, the employed gradient constraints based on the stored pattern information. In the absence of explicit task labels this seems to be challenging. In light of this result, it is particularly interesting to observe that a more straightforward exemplar replay implementation, where the same amount of retained data instances is simply interleaved directly into the continuous training process, seems to achieve accuracies that are sufficiently close to the upper-bound. Naturally, such an approach to continual learning could perhaps be viewed as the most trivial solution, where performance is directly proportional to the amount of retained original data instances. 
	
	\subsubsection{Continual Learning with Quasi-illumination Invariants}
	Following the results of figure \ref{figure:CL_results}, we observe that some deep continual methods are less generically suitable than initially advocated. In practice however, we desire robustness to broader amounts of scenarios, made accesible through our simulator. As an example, mitigating performance degradation as a result of homogeneous lighting changes in the raw video through transformation into illumination invariant spaces has been well known for multiple decades \cite{Nagao1998, Gevers1999, Narasimhan2003}. 
	To showcase the severity of the deep methods' shortcomings, 
	we repeat the naive continuous training in the progressive lighting
	experiment with an included photometric color invariance operation.
	
	\begin{wraptable}{r}{0.425\textwidth}
		\vspace*{-1.25em}
		\captionof{table}{Incremental lighting experiment under consideration of a photometric color invariant or local binary patterns (LBP). \label{table:illumination_invariance}}
		\resizebox{0.36 \textwidth}{!}{%
			\begin{tabular}{cccc}
				& \multicolumn{3}{c}{\textbf{Accuracy} [\%]}\\
				\cmidrule{2-4}
				\textbf{Illumination} & Naive & Naive + & Naive +\\
				\textbf{Intensity} &  & photometric & LBP\\ 
				{[}Lux{]} & & color invariant & \\
				\toprule
				76.8 & 99.20 & 98.66 & 99.18 \\
				& $\scriptstyle\pm^{0.1}_{0.1}$ & $\scriptstyle\pm^{0.15}_{0.19}$ & $\scriptstyle\pm^{0.06}_{0.05}$\\
				\cmidrule{2-4}
				19.2 & 97.11 & 98.61 & 99.27 \\
				& $\scriptstyle\pm^{1.20}_{1.46}$ & $\scriptstyle\pm^{0.47}_{0.98}$ & $\scriptstyle\pm^{0.12}_{0.09}$\\
				\cmidrule{2-4}
				9.6 & 93.55 & 98.61 & 99.26 \\
				& $\scriptstyle\pm^{2.58}_{2.7}$ & $\scriptstyle\pm^{0.21}_{0.36}$ & $\scriptstyle\pm^{0.07}_{0.05}$\\
				\cmidrule{2-4}
				2.4 & 91.55 & 97.56 & 99.42 \\
				& $\scriptstyle\pm^{1.00}_{0.14}$ & $\scriptstyle\pm^{0.76}_{0.76}$ & $\scriptstyle\pm^{0.05}_{0.03}$\\
				\cmidrule{2-4}
				1.2 & 90.89 & 95.28 & 99.40 \\
				& $\scriptstyle\pm^{1.61}_{2.39}$ & $\scriptstyle\pm^{1.32}_{2.07}$ & $\scriptstyle\pm^{0.04}_{0.04}$\\
				\bottomrule
			\end{tabular}%
		}
	\end{wraptable}
	Based on the assumption that color ratios are quasi-invariant under a dichromatic reflection model with white illumination \cite{Gevers1999}, we can define: $c_1 = \arctan (R / \mathtt{max} \{ G, B\} )$, and corresponding definitions for the other two channels. 
	Table \ref{table:illumination_invariance} shows that such preprocessing halves the gap to the upper-bound. Note how inclusion of such a simple assumption already leads to a naive greedy deep network rivalling and even surpassing the accuracies of the continual learning specific designs. Making use of another long known invariant visual descriptor, local binary patterns \cite{He1990,Wang1990,Ojala1994,Ojala1996}, the accuracy in table \ref{table:illumination_invariance} even closely approaches a 100\%, see the appendix for further details. This further highlights the importance of considering the nature of diverse scenarios, as deep continual learning should ideally leverage quasi-invariant spaces where possible to be stable. 
	
	\section{Discussion of Simulator Use-Cases and Prospects for Analysis}
	Our presented empirical investigation has consciously focused on rather simple classification tasks in an attempt to provide an initial experimental showcase for our simulator's utility. The rationale behind this choice has been two-fold: a) The experiments should be directly relatable to the community with respect to following the predominant set-ups of prior investigations, e.g. incremental MNIST, CIFAR and similar continual classification practices (even though it could be argued whether seeing only trees, or only cars is a realistic assumption in practice). b) The experiments seem to sufficiently demonstrate that the phenomenon of catastrophic interference in continual learning requires a more principled exploration in more diverse and controlled settings. 
	
	We note that, as a benefit of our flexible simulator design and its accessible modularly parametrized generative model, the presented classification experiments represent but a small subset of readily assessable future experiments. To point out the present investigation's limitations, we highlight immediately conceivable investigations in a short outline:  
	
	\begin{itemize}
		\item \textbf{Semantic segmentation \& modalities:} Continual learning investigation in semantic segmentation \cite{Ronneberger2015,Eigen2015,Kendall2017, Kohl2018}. On the one hand, we can conduct experiments in direct extension to the presented classification ones. On the other hand, we could investigate an alternative where objects are always present and instead labels are progressively added to become increasingly fine-grained, see the example of figure \ref{figure:render_examples}. Similarly, analysis can be extended through consideration of the simulator's other modalities, such as surface normals and depth.  
		\item \textbf{Frequency of occurrence:} Our experiments have presently focused on introduction of a single class at a time or illumination and weather conditions being equally likely. In practice, it is certainly the case that probability of occurrence plays a major role. We expect that future investigations can adapt our presently assumed probabilities, for instance, in order to investigate scenarios with rare occurrences, or continual learning scenarios where concepts appear or disappear multiple times throughout the entire video sequence. 
		\item \textbf{Identifying distribution shift:} Present continual learning mechanisms of the experiments were provided with task-boundary information. The increasingly important question of whether a devised approach can identify various sources for, potentially continuous or gradual, distribution shift can and should be considered (e.g. to decide when to learn continually or when to protect the model from catastrophic interference) . 
		\item \textbf{Disentangled representations:} Apart from above straightforward prospects, recall that our simulation has explicitly laid open and parametrized physics-based rendering properties. This leads to multiple imaginable video sequences that we believe will have particular importance for future work. In figure \ref{figure:render_examples} an example where the scene is rendered without material object normals, surface roughness and is devoid of color has already been depicted. There is no light reflections or refractions such that the image presents a simplified gray-scale world with a focus on geometry. Governing such physics-based rendering properties in conjunction with real-time control over appearing objects and environmental conditions can facilitate future analysis into the disentanglement of representations in deep generative models \cite{Achille2018, Higgins2017, Mathieu2019}, enable further investigation into the debate on texture versus shape bias in deep learning \cite{Ritter2017,Geirhos2019}, or allow for the analysis of meaningful learning curricula of increasing complexity \cite{Bengio2009}.
		\item \textbf{Temporal consistency:} Finally, it is worth to remember that our simulator renders temporally cohesive video streams, even though our initial classification experiments have considered frames independently. All of the above suggestions can thus be conducted under consideration of temporal consistency, arguably being a natural mode of data presentation for continually learning systems.
	\end{itemize}
	
	In addition to these mentioned prospects, we point out that future analysis could also consider the degree of transfer from trained simulated models to the real-world. Although this is not the essential premise of our work, the latter could be regarded as a present limitation of our work. As with any simulator, the degree of transfer inevitably scales with the availability of high-fidelity assets.
	To this end, please see the appendix, where we provide a more detailed discussion on this issue with respect to our simulator.
	Nevertheless, we recall that catastrophic interference should ideally first be overcome in well understood simulation before deducing generic mechanisms with opaque interpretations on real-world applications. A final assimilation of simulator to real-world statistics could thus be regarded as a subsequent goal \cite{Veeravasarapu2017, Shrivastava2017}. 
	
	\section{Conclusion}
	We have introduced a parametric interpretable generative model and its 3D graphics engine realization for the procedural online generation of continual learning scenarios. 
	It provides a rich set of flexible generative factors that are adjustable by a straight forward configuration and cover all aspects of the continuously evolving virtual world. This allows the user to easily generate temporally consistent data streams which would require potentially insurmountable effort to be acquired in the real-world.
	To bootstrap the proposed benchmark generator, initial exploration on the basis of three distinct generated scenarios has been aligned to the currently employed evaluation scheme of using a set of successive tasks, each composed of an iid classification dataset. 
	Without raising the complexity to temporally consistent online learning, or objectives such as semantic segmentation, the presented experiments already highlight the necessity of such a simulation for more extensive evaluation, in order to analyze and overcome the current shortcomings of continual deep learning mechanisms.  
	
	\begin{minipage}{0.25\textwidth}
		\includegraphics[width=1.2\textwidth]{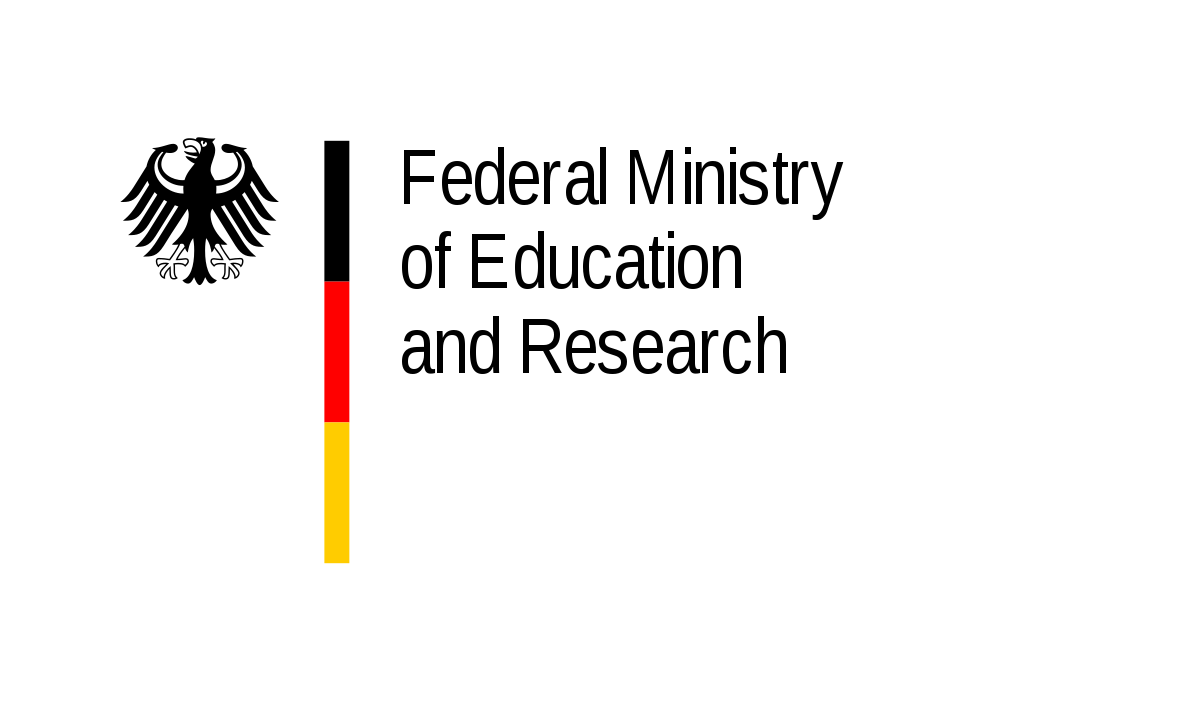}
	\end{minipage}
	\begin{minipage}{0.75\textwidth}
		\textbf{Acknowledgements:} This work was supported by the Artificial Intelligence Systems Engineering Laboratory (AISEL) project under funding number 01IS19062, funded by the German Federal Ministry of Education and Research (BMBF) program "Einrichtung von KI-Laboren zur Qualifizierung im Rahmen von Forschungsvorhaben im Gebiet der Künstlichen Intelligenz".
	\end{minipage}
	
	\clearpage
	
	{\small
		\bibliographystyle{unsrt}
		\bibliography{references}
	}
	
	\clearpage

\appendix

\section*{Supplementary Material: Content Overview}
The supplementary material contains further details for the content presented in the main body. The first two sections provide additional information on the simulator, spanning its 3D assets and configuration interface for the user, as well as its limitations. Thereafter, datasets generated for the classification tasks are discussed in more detail, and  
we elaborate on our conducted experimental training details, hyper-parameters and used neural network architectures.  
In the ultimate section of the supplementary material, we provide a description and intuition behind the photometric color invariance operators, as used in the main body to demonstrate the severity of the shortcomings of the deep continual learning algorithms in the progressive lighting experiments. In summary, the overall structure of the supplementary material is as follows:

\begin{itemize}
	\item[\textbf{A.}] Additional simulator details
	\item[\textbf{B.}] Limitations of the simulator
	\item[\textbf{C.}] Dataset generation for classification
	\item[\textbf{D.}] Deep neural network training hyper-parameters
	\item[\textbf{E.}] Operators for quasi-invariance to illumination
\end{itemize}

Please note that further detailed readme files, containing extra instructions on installation for our software contributions, as pointed out in the main body, can be found in the respectively linked repositories.

\section{Additional Simulator Details} \label{section:simulator_details}
In section 2 of the main body we have introduced our key contribution of a computer graphics simulation framework for the flexible composition of data streams that facilitate assessment of continual learning. There, we have detailed the underlying modular generative model and its random variables. Recall, that a majority of objects and actors are placed stochastically and their appearance is modelled largely through categorical variables. In this supplementary section, we further illustrate the currently available 3-D assets that are subject to this categorical sampling, their placement in the scene, and the overall possibilities of configuration for the simulator to re-emphasize its flexibility in creating specifically customized scenarios in the context of continual learning.  
\begin{figure}[h]
	\centering
	\includegraphics[width=0.12\textwidth, height=0.125\textwidth]{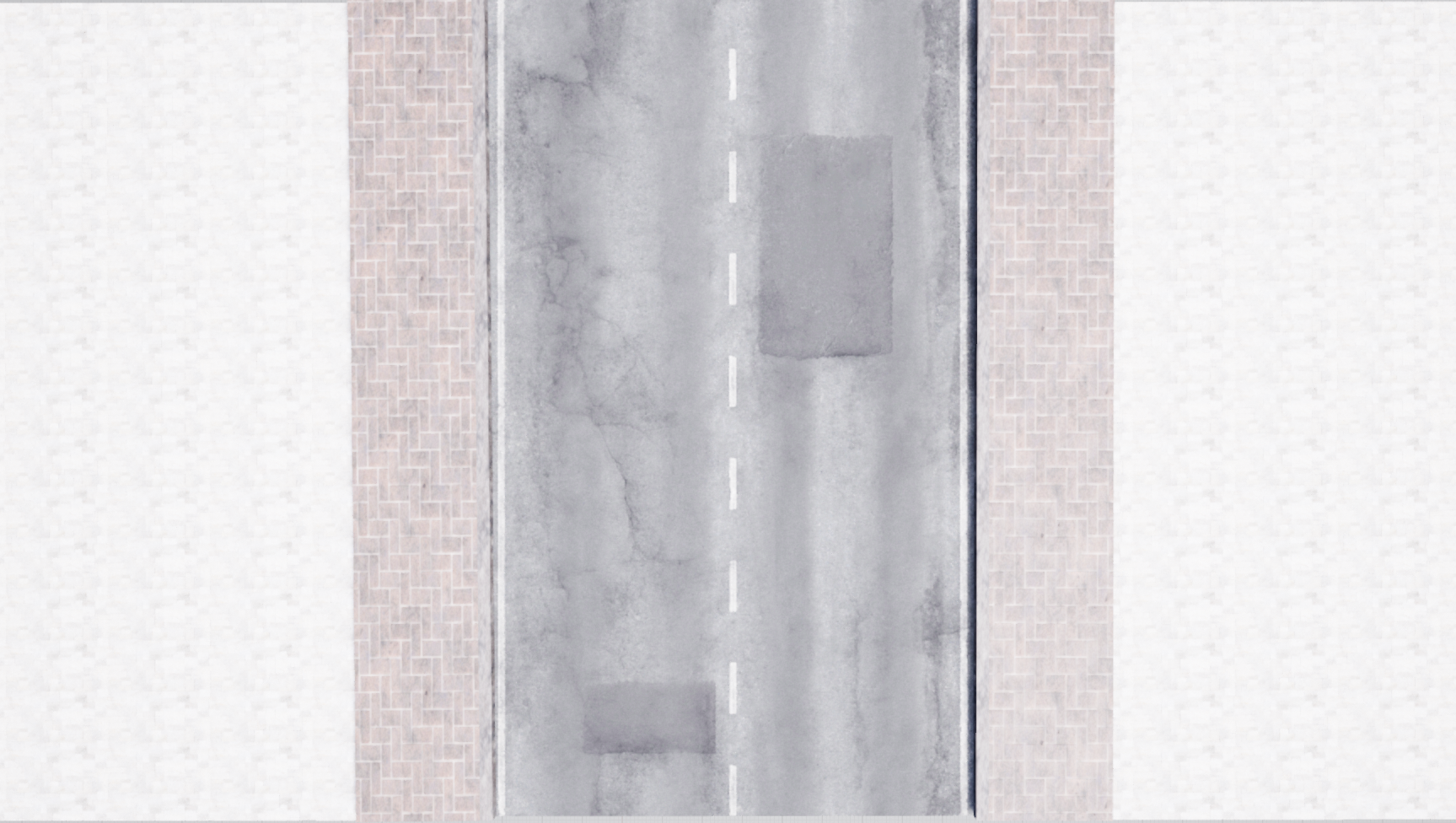} \, \includegraphics[width=0.12\textwidth]{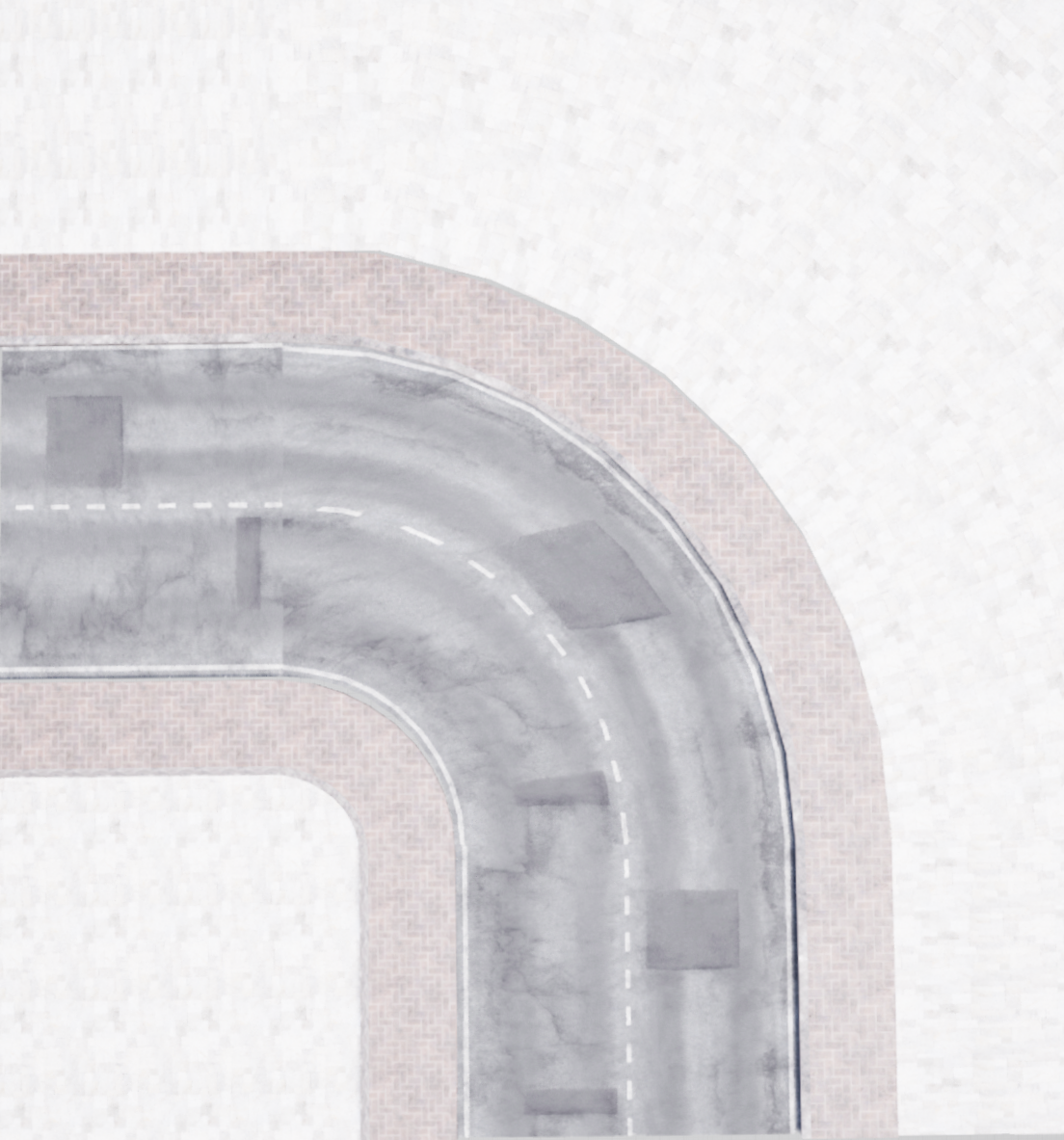} \,
	\includegraphics[width=0.12\textwidth]{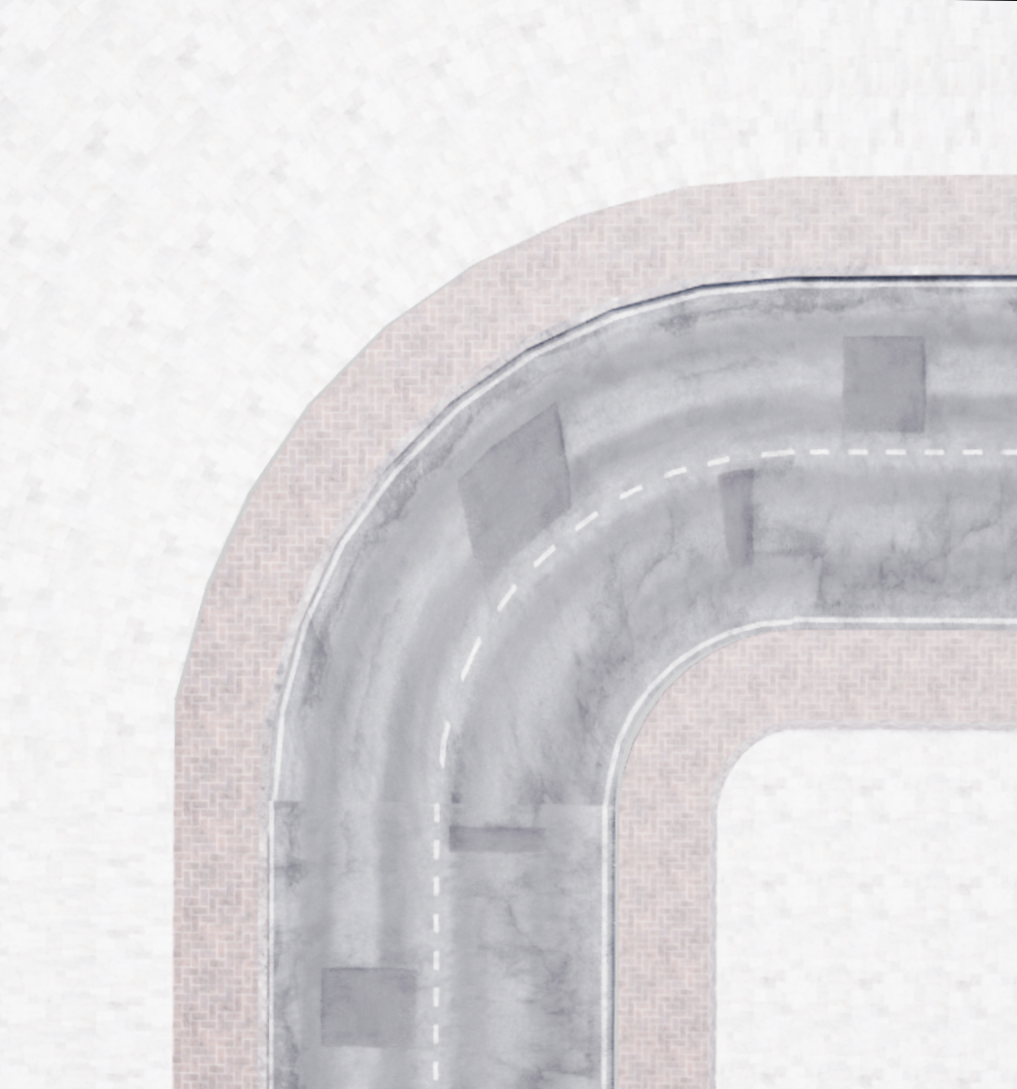} \,
	\includegraphics[width=0.54\textwidth]{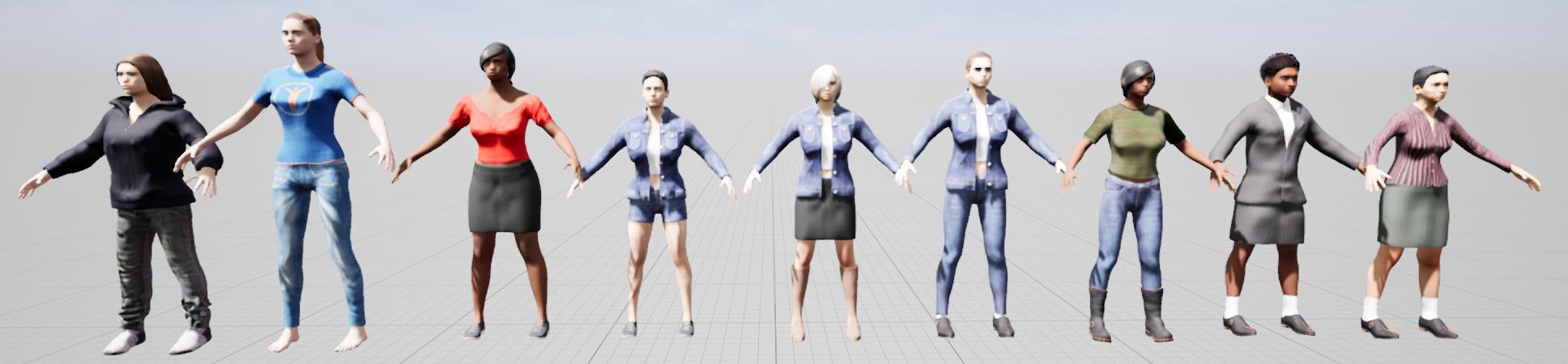}\\
	\ \\[-0.8em]
	\includegraphics[width=0.12\textwidth]{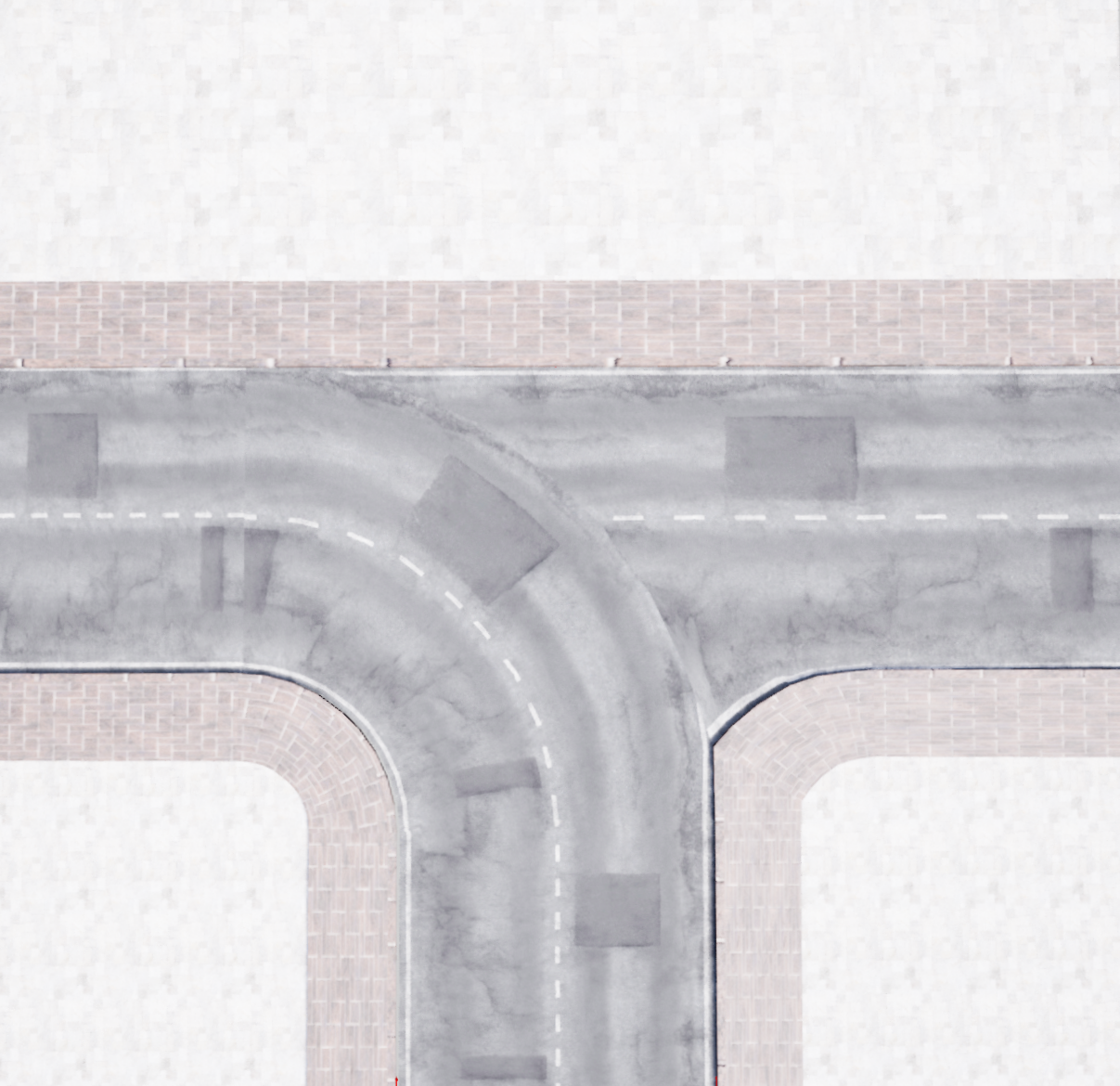} \, \includegraphics[width=0.12\textwidth]{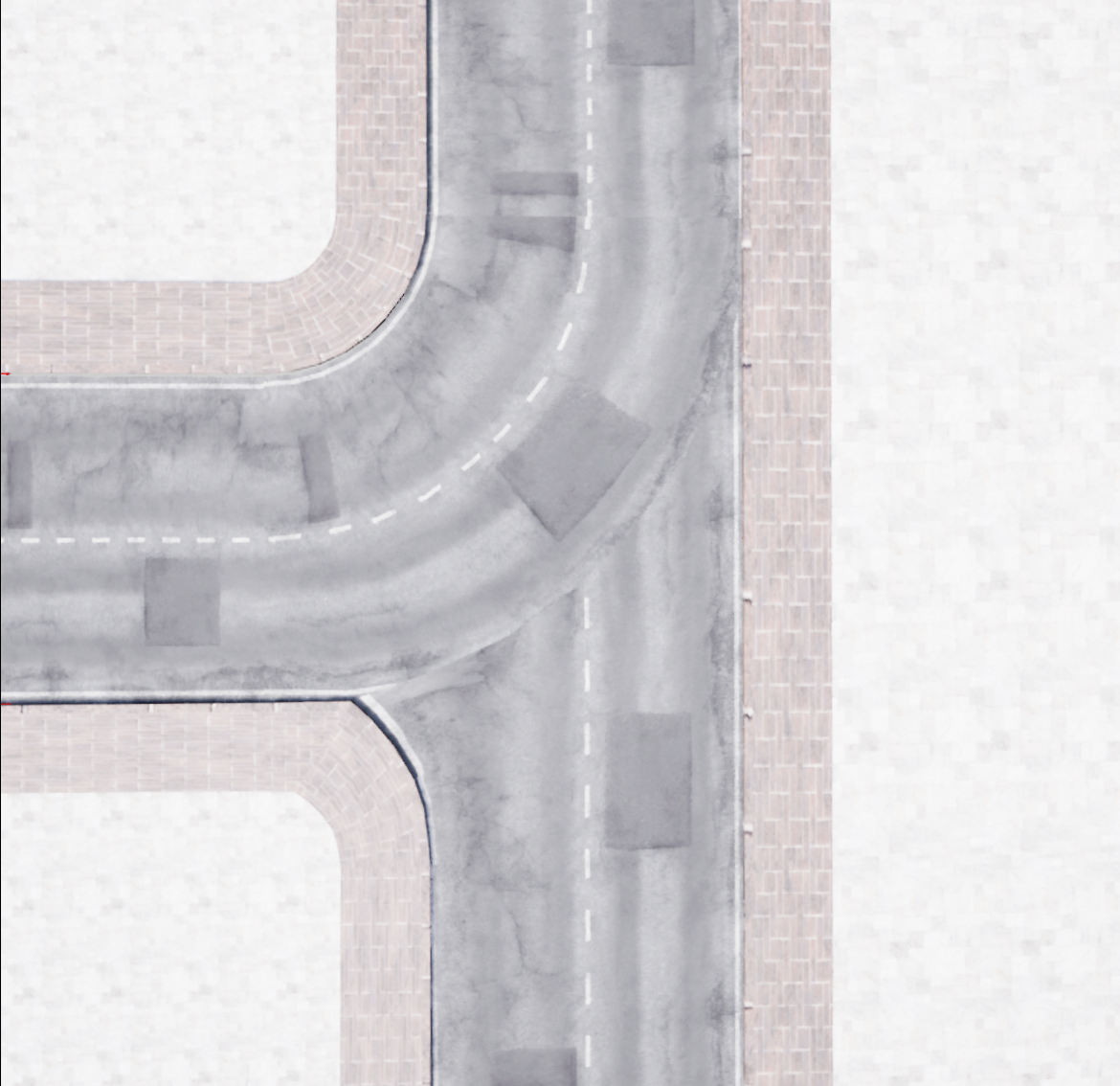} \,
	\includegraphics[width=0.12\textwidth]{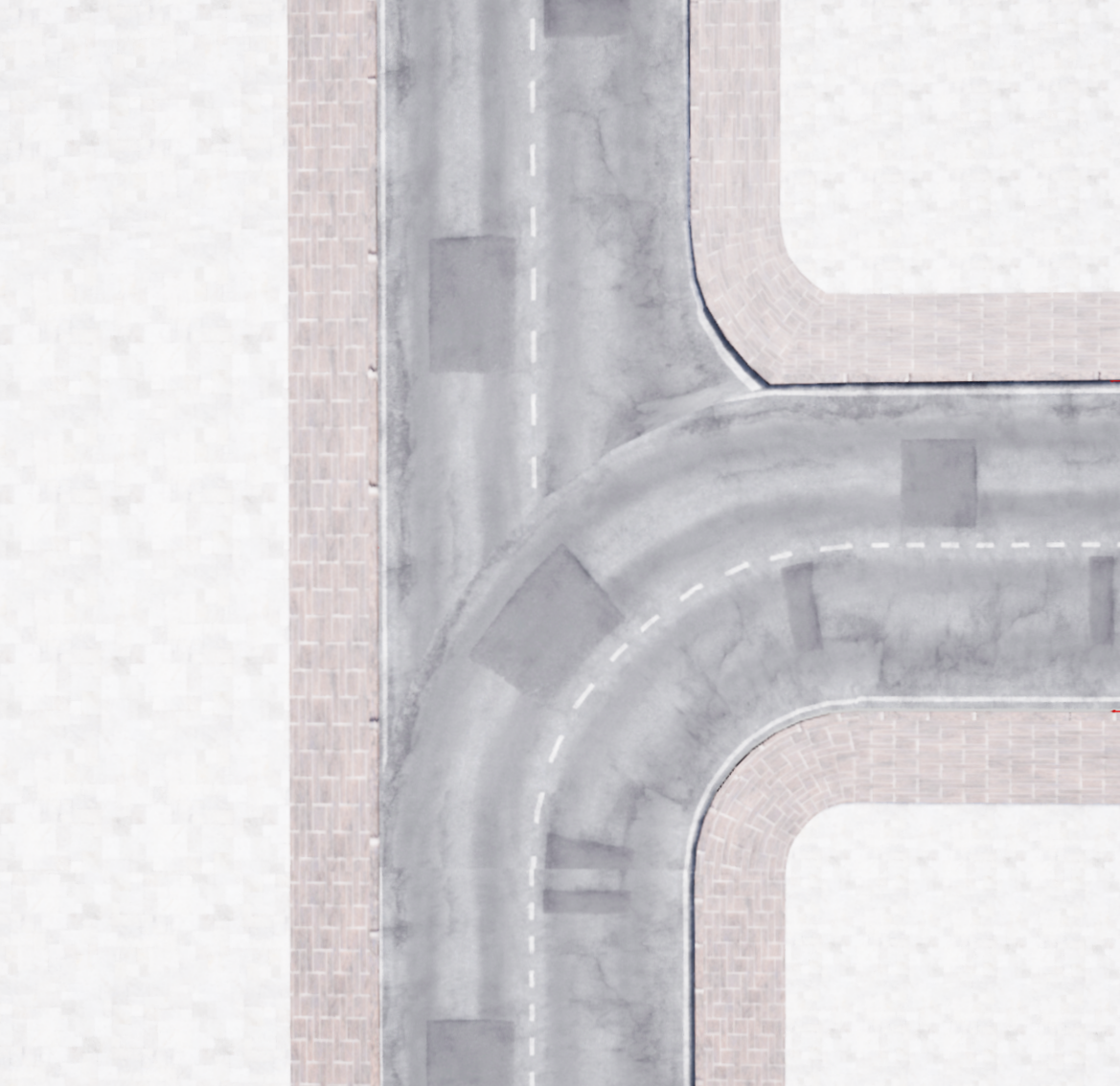} \,
	\includegraphics[width=0.54\textwidth]{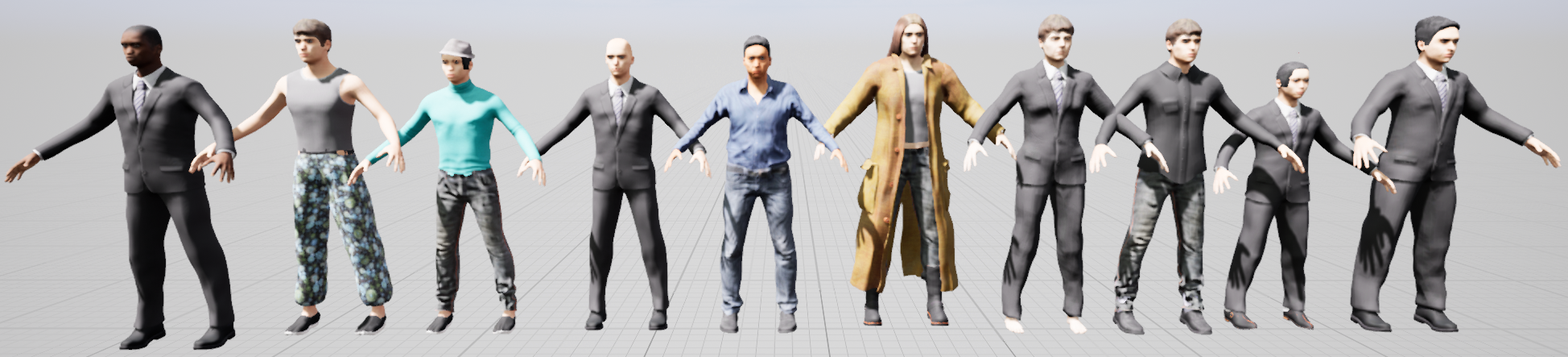}\\
	\ \\[-0.8em]
	\includegraphics[width=0.43\textwidth]{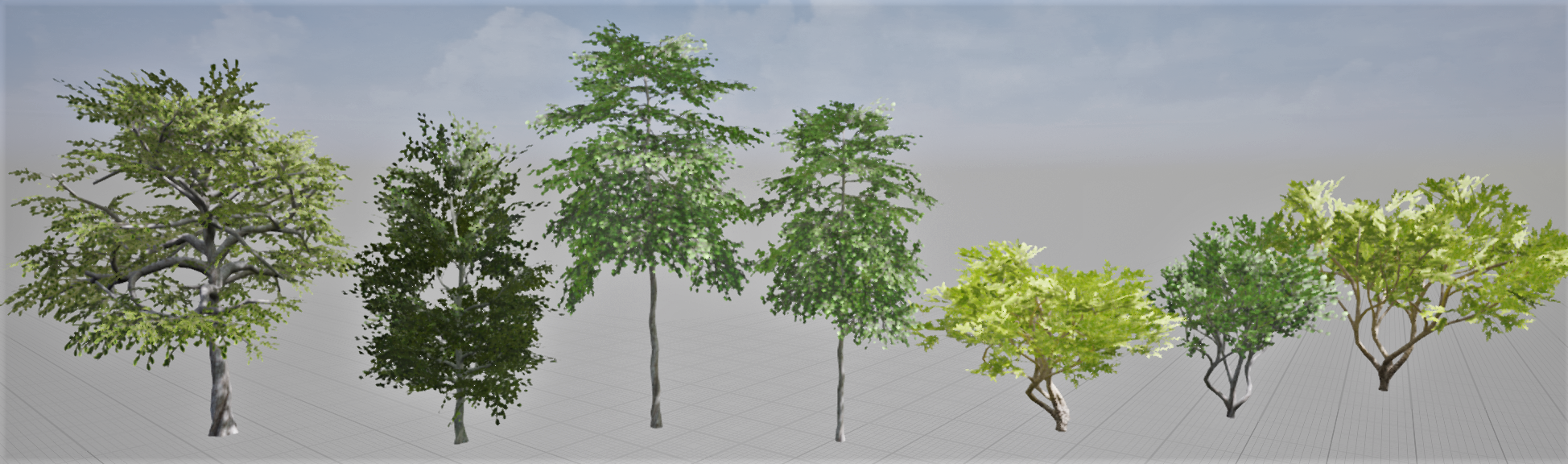}
	\includegraphics[width=0.04\textwidth, height=0.127\textwidth]{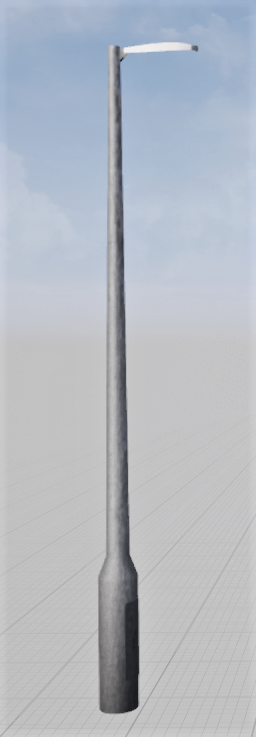}
	\includegraphics[width=0.47\textwidth, height=0.127\textwidth]{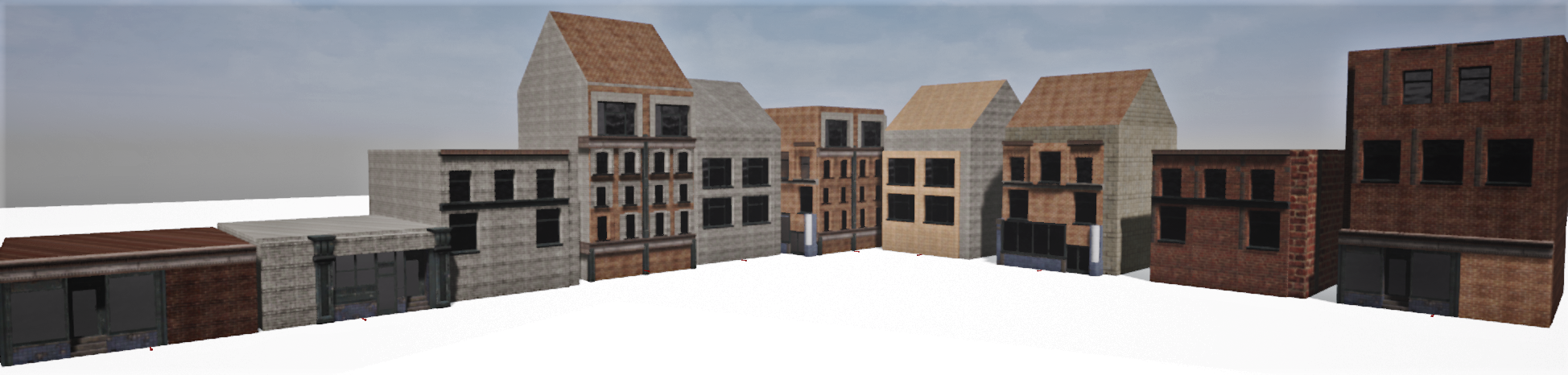}\\
	\ \\[-0.8em]
	\includegraphics[width=0.8\textwidth]{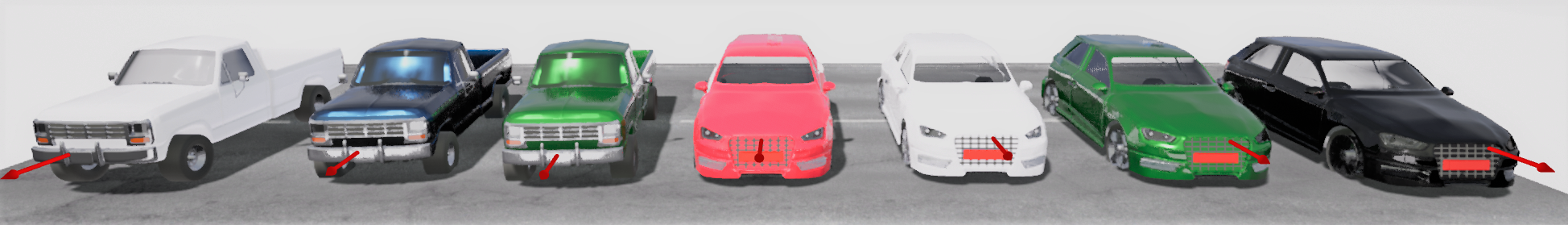}
	\captionof{figure}{Overview of 3-D assets used in the current implementation of our simulator, spanning 6 street segment layouts, 10 buildings, 19 human actors, 7 trees, 1 street lamp, and 2 cars with corresponding 3 and 4 colorings.
		\label{figure: object selection}}
\end{figure}

\textbf{Sampled assets:} In the main body, we have already described $E_{i}$, a convenience variable which defines whether a certain object or actor category i, for example a tree or a human, is sampled into the generated scene. Depending on the result or the user's choice, a spawned object or actor is then subject to categorical sampling, followed by stochastic placement on a street tile. This categorical sampling represents a simple process of randomly selecting a 3-D asset from the available repository. Recall, that for each street tile this categorical sampling is repeated in order to place multiple objects or actors of the same type within the bounds of each randomly selected street segment. As such, controlling the probabilities on the categorical variables and extending or limiting the used repertoire of 3-D assets directly impacts the diversity of possibly generated scenes. 

Figure \ref{figure: object selection} illustrates the assets that have currently been included in our simulator for the presented experiments. The figure shows the six available street segment types, representing right and left curves, a continuing straight road and three choices for crossings. The discretized set of buildings and trees can be observed to have been picked to balance variations in width and height. Vehicles vary in color and car model. Pedestrians were selected with different body type, skin color, age and gender in mind. The presently used amount of distinct 3-D assets is made up of 6 street segment layouts, 10 buildings, 7 trees, one street lamp, 2 cars with respective 3 and 4 colors, and 19 human actors. These assets where mostly generated by hand, using editors such as \textit{Blender}\cite{BlenderFoundation}, \textit{TreeIt}\cite{EVOLVEDSoftware}, \textit{MakeHuman}\cite{MakeHumanCommunity} and modular asset packs provided by the Unreal Engine store \cite{EpicGames2015}.

The heterogeneity of the scenes is thus presently controlled solely through the set of available assets and their categorical sampling probabilities. In order to allow for more nuanced control, it is however imaginable to further decompose these categorical variables into separate complementary variables that manage e.g. the geometry of a car and its color independently. 
For our current investigative CL experiment purposes, the purely categorical sampling has been observed to be sufficient and further extensions are left to future work.

\begin{figure}[t]
	\begin{tabular}{p{0.1\textwidth} p{0.115\textwidth} p{0.10\textwidth} p{0.115\textwidth} p{0.06\textwidth}}
		\hspace*{0.1em} \scriptsize Terrain & \scriptsize Sidewalk & \hspace*{0.5em}\scriptsize Street & \scriptsize Sidewalk & \scriptsize Terrain\\[0.3em]
		& & & &\\[-2em]
	\end{tabular}
	\centering
	\includegraphics[width=0.7\textwidth]{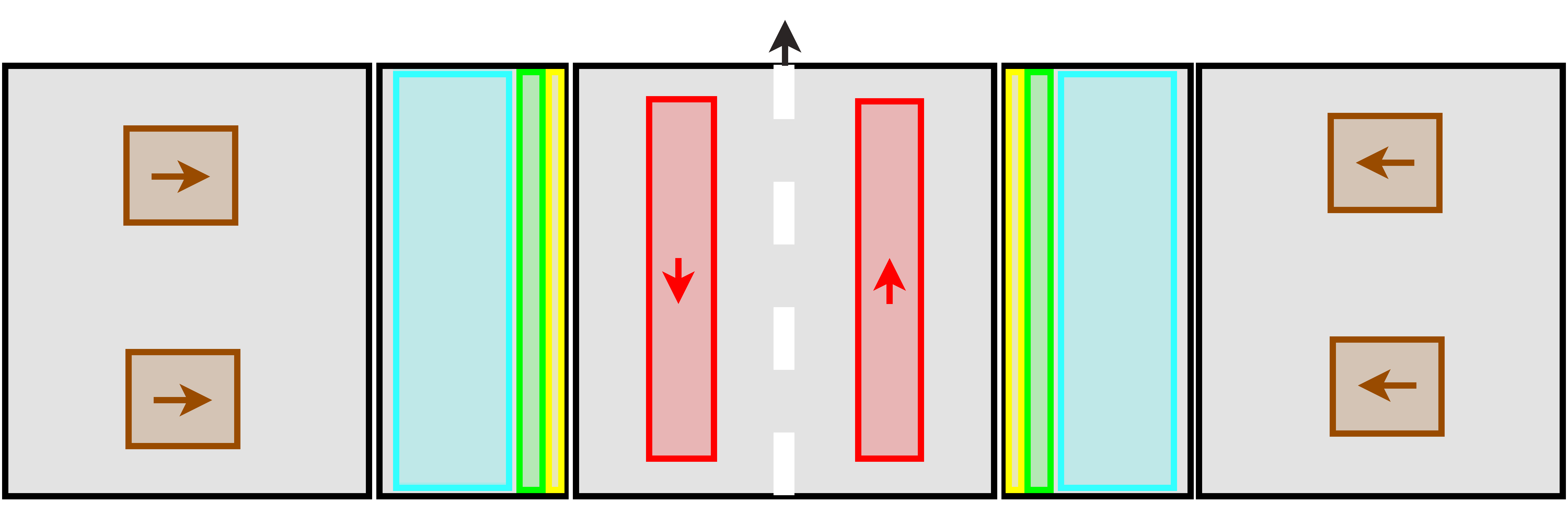}
	\captionof{figure}{Schematic illustration of the object placement for the example of a straight street segment. The bounded areas for each object category are color coded as follows: vehicles (red), streetlamps (yellow), trees (green), pedestrians (cyan), buildings (brown). Arrow-markers inside these areas indicate an additional direction specification in which spawning objects should face. This way, vehicles do not drive against their lane direction, and building-fronts face the street. The black arrow-marker at the top of the tile, indicates the spawn anchor for the next sampled tile's attachment. \label{figure:tile_structure}}
\end{figure}

\textbf{Sampled positioning:} Sampling the position of an object upon placing it into the scene is implemented by a finite support uniform distribution, with its bounds resembling intuitive assumptions on real-world statistics, such as trees not being planted in the middle of a road. In figure \ref{figure:tile_structure} we show a schematic illustration of the bounds for each object category's placement for the example of a straight street segment. Object-spawning areas are depicted as colored bounding boxes, which indicate the limits for the uniform sampling of each object's placement in the tile's space with respect to its category. These areas are individually defined for each tile layout, however, all bounds' definitions follow the same structure of design, as is illustrated in the figure. Providing each object category its own, non-overlapping, space on the tile is a deliberate choice to reduce the complexity to prevent objects of different categories from overlapping or otherwise colliding at runtime.
Collision between objects of the same category, e.g. humans being placed into one-another, are initially avoided by the spawn routine itself checking for overlaps when placing an object into the scene. For dynamic actors collision avoidance is handled by their behavioral implementation.

\textbf{Configuration:} 
Simulator configuration by the user is a straightforward process on the basis of the generative model for video sub-sequences, as defined in section 2.1 of the main body. Being able to manipulate the parameters of this model, the user is provided with the freedom to define arbitrary chains of sub-sequences in the generated video stream. We re-emphasize that this translates to a vast number of scenarios to be generated with nuanced control over abrupt  and/or continuous distribution changes. 
In the executable version of the simulator, provided with this work, all configuration of the sequences to be constructed can be specified prior to actually running the data generation process. The configuration itself is interfaced by JSON-config files. Example files containing the configurations to the data generation for all experimentation conducted in the main body are part of the repository containing our simulator.

\begin{figure}[b]
	\centering
	\includegraphics[height=0.275\textheight]{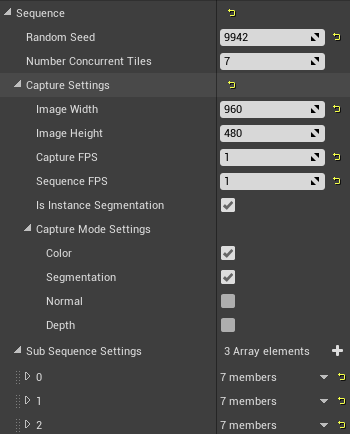} \,
	\includegraphics[height=0.275\textheight]{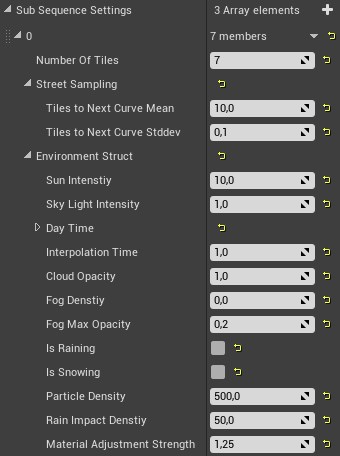} \,
	\includegraphics[height=0.275\textheight]{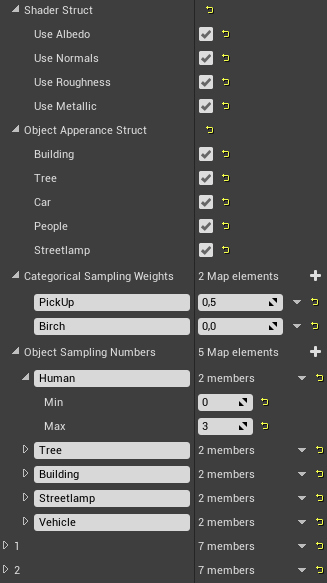}
	\captionof{figure}{Overview of a sequence's parameter configuration as visualized by the Unreal Engine editor. Left: parameter settings that are applied to the entire stream, including rendering options and random-seed. Center and right: parameters settings of an example individual sub-sequence, outlining the range of configuration possibilities. \label{figure:simulator_configuration}}
\end{figure}

For the following further explanation of the configuration parameters we proceed with the Unreal Engine editor's view of a sequence's configuration. A corresponding visualization is depicted in figure \ref{figure:simulator_configuration}. We note that the presented visualization of a sequence configuration is automatically generated by the Engine's-Editor from our data-structure holding the configured parameters. 

Overall, the configuration of a sequence can be understood as being divided into two major parts.\\ 
The first part defines parameters that are applied to the entire stream. These stay unchanged for all sub-sequences and include the camera's settings $\theta_C$, such as the frames-per-second and resolution of the rendered output and the render-modes to be considered. Additional included parameters are a general random-seed that is the basis for all sampling in the sequence, and the number of tiles to make up the main street track at any time during the simulation. \\
The second part covers all remaining detailed parameters to provide the specification for each individual sub-sequences. The length of a sub-sequence is specified by the 'Number of Tiles` to be sampled. The 'Street Sampling` covers $\theta_S$, controlling the curvature in the street track. The 'Environment Settings` are composed of $\theta_L$ and $\theta_W$, manipulating lighting and weather related variables. The 'Material Settings` control the shading options, equivalent to $\theta_R$. The 'Object Presence` governs parameters $\theta_E$ for the Bernoulli variables.\\ 
Finally, the 'Categorical Sampling Weights` used for the conditional selection of specific objects can be adjusted, which correspond to $\theta_H$, $\theta_V$, $\theta_{Tr}$, $\theta{Lp}$, $\theta_B$. Similarly,  'Object Sampling Numbers` control the number of said objects $N_H,N_V,N_{TR},N_{Lp},N_B$ to be sampled for each tile.
The categorical sampling weights are specified by mapping identifiers, i.e. string-typed names, to probability values. These correspond to names in the directory structure, which is used to store the respective object-assets inside the simulator. This way, sampling probabilities can be assigned to individual object instances, e.g. one specific birch tree, or alternatively, to entire sub-categories, e.g. all birches. The exact directory structure is provided to the user as part of the repository of this work.\\
A similar specification process is used for the object numbers. However, here only top level identifiers, that correspond to respective object categories, e.g. tree, car, or people, are considered. 

To lift potential confusion regarding the interplay of 'Object Presence` settings (parameters of $\theta_E$), 'Object Sampling Numbers`, and 'Categorical Weights`, we briefly recapitulate their relations: as already touched upon in the main body, the 'Object Presence' settings are convenience parameters, to easily allow or deny sampling of certain object categories. This design has been chosen in light of often applied incremental class continual learning scenarios. The same effect could be achieved by setting 'Object Sampling Numbers` for the respective category to a value of 0. However, once the 'Object Presence` for a given object is turned off, further 'Object Sampling Numbers` are ignored. Regarding the categorical weights in this context, we want to highlight that these are understood as being conditioned on the event of an object being sampled. Setting an object-category identifier to zero, or likewise all identifiers of this particular category to zero individually, will fall back to consistently sampling the first asset entry of that category.  

\section{Simulator Limitations} 
Building on the previously described extended information on assets and sequence configuration, we discuss the limitations of the provided simulator in its current state. 
As we pointed out, the heterogeneity of the scenes to be generated is eventually dependent on the diversity of the 3-D assets available for sampling. This is because the restriction to the used asset repository naturally affects the number of achievable, truly distinct, tile configurations. In principle, the data generation may 'endlessly' sample a continuous world, which is in contrast to static maps that can be internalized over time. However, there is inevitably going to be a re-occurrence of some previously observed patterns after the repertoire has been exhausted in all possible configurations (which we re-emphasize is still a large number considering randomness in locations, environmental factors etc.)
Towards the ultimate goal of transferring models trained in virtual environments to applications in the real-world, which is the most common use-case of synthetic data generation, the total number of distinct possible patterns to be observed should be maximized and likewise, the distance to matching real-world statistics minimized. 
This issue is presently inherent to any graphics simulation for computer vision and has seen multiple solution attempts, as elaborated in the main body's related work section. At this point, adaptation from our simulator to real remains a desideratum. 
Even though this limitation is recognized, we re-emphasize that our main body's experimentation suggests that there nevertheless is large merit in the generated datasets as benchmarks for continual learning.

One imaginable way to easily overcome the asset limitation is through the inclusion of further publicly provided assets. Online repositories for the latter are fortunately growing at a rapid pace. However, 
being publicly accessibly and both non-commercially and commercially distributable are two distinct factors. Most commonly used licensing unfortunately prohibits re-distribution by any third-party, particularly if assets require a prior purchase. This includes some of the above mentioned assets, where we have referenced third party software respectively. For our open-source contribution we have thus opted for a two-fold distribution of our work: the source code of our simulator, where various assets have been removed, but the core code mechanisms are accessible and can be build upon. We believe that reintegration of assets and inclusion of new ones can be managed with reasonable effort. At the same time, for the continual learning practitioner and investigator, we also distribute our simulator in a self-contained and encrypted executable, as encouraged by Unreal Engine's market place policy. Although the assets cannot exchanged in this version, it provides the user with the necessary means to set the generative process' parameters through configuration files, as described in the previous section.

\section{Dataset Generation for Classification} \label{section:data_generation}
An empirical analysis of multiple continual learning scenarios has been shown in the experimental section of the main body. There, we have argued that the simplest conceivable assessment is a classification setup. The respective empirical results have indicated that this already provides a significant challenge in consideration of the even more complex semantic segmentation, surface normal or depth prediction tasks. In this supplementary material section we first provide additional detail on general parameter choices for the generation of our video streams, illustrate how they can almost trivially be used for classification purposes, and then proceed to further specifying the precisely constructed datasets.
Before continuing, we would at this point like to remind the reader that the constructed classification scenarios present but a tiny fraction of the imaginable applications of our proposed simulation framework, as thoroughly explained in the main body's discussion. 

\textbf{Parameter Choices:} For our current purposes, the default parameters of all categorical variables in the generative model are set such that each of the $M$ choices is equally likely: 
\begin{equation}\label{eq:object_probabilities}
	P(Obj = Obj_m \, | \,  \theta_{Obj}) = 1/M \quad \forall \, m = 1, \ldots, M \, .
\end{equation}
The parameters governing the number of individual objects and actors is set in analogous fashion, with empirically set maximum number of 4 buildings, 8 humans, 2 additional cars, 4 street lamps and 6 trees per street segment. The number of tiles $N_S$ is treated as deterministic and set to 7 at a time, as a practical trade-off to surpass the view distance of the employed camera. 
In order to prioritize multiple straight street segments before occurrence of a curve or crossing, the number of the next consecutive straight street segments is predetermined by drawing from a normal distribution $n_S \sim \mathcal{N}(4, 0.45)$ each time a straight street is spawned. The specific straight street layout for the individual segments is then determined by:
\begin{align}
	P(S = Straight_i \ | \theta_S) = (1/I), \forall i = 1,...,I \, .
\end{align}
After $n_S$ straights have been placed, a curve or crossing is spawned with equal sampling weights for all available layouts, resulting in their placement probabilities:
\begin{align}
	& P(S = Curve_k \ | \ \theta_S) = (1/K) \cdot 1/2 \, \, \, \forall \, k = 1, \ldots, K\\ \nonumber
	& P(S = Crossing_l \ | \ \theta_S) = (1/L) \cdot  1/2 \, \, \, \forall \, l = 1, \ldots, L \,\,  ,
\end{align}

where I, K, L indicate the respective number of available distinct street tiles. For an initial practical assessment of continual learning, only one straight tile of constant width and length, two choices for left and right turning curves and three variations for crossings have been included. 
We have visualized the collection of currently available object and actors in previous section's figure \ref{figure: object selection}.
These parameter settings are kept constant over time. In contrast, considered continuous learning scenarios can be characterized through a change of parameters over time. 
Whereas this could be nuanced modifications in above probabilities on specific object models or locations over time, we choose to investigate sequences with respect to variations in weather, lighting and the Bernoulli variables $E$ for existence of entire object categories, mirroring class incremental scenarios. 
As an example, an incremental weather video sequence can be generated by initially sampling the categorical weather with equal probability, akin to equation \ref{eq:object_probabilities}, and subsequently setting the already sampled condition's probability to zero and raising the remaining choices' likelihoods to sum to unity. Alternatively, the probability for a desired categorical outcome can just be set to one. In the same spirit the existence of particular objects defined through Bernoulli variables $E_B, E_{Tr}, E_{Lp}, E_H, E_V$ can just be assigned probability vectors. For instance, short notation $\pi_{E, t=0} = (1, 0, 1, 0, 0)$ indicates the presence of buildings and lamps, and the absence of humans, additional vehicles and trees in a video sub-sequence. Exact considered set-ups are described in the experimental section.

\newpage 

\begin{wrapfigure}{r}{0.475\textwidth}
	\centering
	\includegraphics[width=0.435\textwidth]{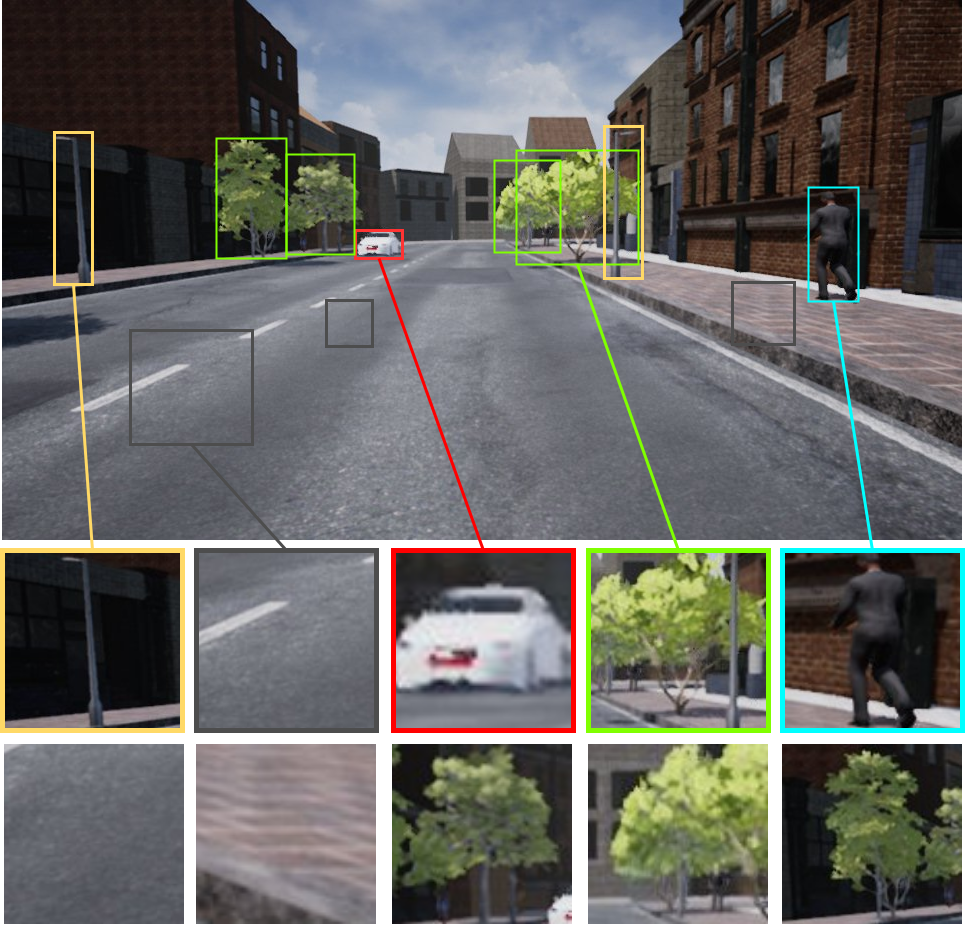}
	\captionof{figure}{Example video frame illustrating the extraction of squared image-patches for a pure classification task. Given respective bounding boxes for each object instance, a quadratic crop is taken based on the larger of the two bounding box dimensions, such that the object is centered in the resulting patch. Example extracted image patches have been connected to their original bounding boxes to provide better intution for this process. \label{figure:converison_example}}
	\vspace*{-1em}
\end{wrapfigure}
\textbf{Classification Dataset Construction:} Given that our simulator already provides exact pixel-wise ground truths of the scene, construction of the classification tasks for our class incremental, incremental weather and incremental lighting scenario is but a matter of straightforward image patch extraction from video frames. In other words, given an object's bounding box only the content enclosed by its extent is presented to the neural network to learn a simple prediction with respect to the contained class. Intuitively, this can be thought of as being granted access to exact object positions in the video stream by an all-knowing oracle. An example video frame with depicted bounding boxes and their extracted contents is illustrated in figure \ref{figure:converison_example}. At a closer look, it can be observed that the bounding boxes are of a general rectangular format, with e.g. a street lamp always being much taller than its corresponding width. In contrast, the extracted image patches for the classification task always featuring a 1:1 aspect ratio. This represents a choice and further simplification in dataset construction, as a direct consequence of assuming a deep convolutional neural network that is designed for static quadratic spatial input dimensions, see supplementary part \ref{section:hyper_parameters} for neural network details. 

We can summarize the explicit extraction procedure from a full video frame to separate square image patches in a few key steps: 
\begin{enumerate}
	\item \textbf{Object bounding boxes:} Given the exact semantic segmentation pixel-wise ground-truths, a tight bounding box is directly acquired based on the furthest labelled pixels with respect to the height and width dimensions. 
	\item \textbf{Background bounding boxes:} For each video frame, four "background" bounding boxes are randomly sampled in order to include regions that are not explicitly included as a separate category in the classification task. To imitate the average extent of the bounding boxes for the other classes, background width and height are randomly sampled from a uniform distribution in a range of $64$ to $200$ pixels. To avoid conflicts between categories of interest and the background, a sampled background bounding box is rejected if it collides with an already existing object bounding box of the particular frame. 
	\item \textbf{Image patch crops:} Create a set of image patches for each video frame by taking crops of the bounding boxes. In our specific case of a later trained neural network that requires spatially symmetrical input, we take a square crop based on the larger of the bounding box's dimensions, such that the contained object is centered. In other words, a small amount of background is included for the shorter side of the bounding box. We do not at this point randomize the cropping to contain stochastic translations of the object, as is commonly done in training of datasets such as ImageNet \cite{Russakovsky2015} for data augmentation purposes. 
	\item \textbf{Consistency step:} For our main body's experiments we have chosen to include a further simplification step to facilitate the classification and assure consistency between extracted patches. Specifically, we discard any bounding boxes and resulting crops with a minimum width or height that does not surpass $32$ pixels. That is we do not include objects that are far in the distance in our current assessment. We also guarantee that the classification task is strictly single target by excluding bounding boxes that have more than 20\% overlap with adjacent objects. That is, we do not include image patches that feature more than one class at its center or contain any occluded objects. 
\end{enumerate}

We further note that for our presently constructed classification dataset, we skip above steps 1-4 whenever subsequent video frames are identical. The latter scenario can arise e.g. when the camera car is stopped at a street intersection. This represents a simple choice to balance the constructed classification dataset as much as possible and avoid heavy redundancy. For all videos the camera model is assumed to be unchanging and corresponds to Unreal Engine 4's default camera, with an effective shutter speed of $1/s$, an ISO value of $100$ and an Aperture equal to unity. 

\begin{wrapfigure}{r}{0.45\textwidth}
	\centering
	\includegraphics[width=0.45\textwidth]{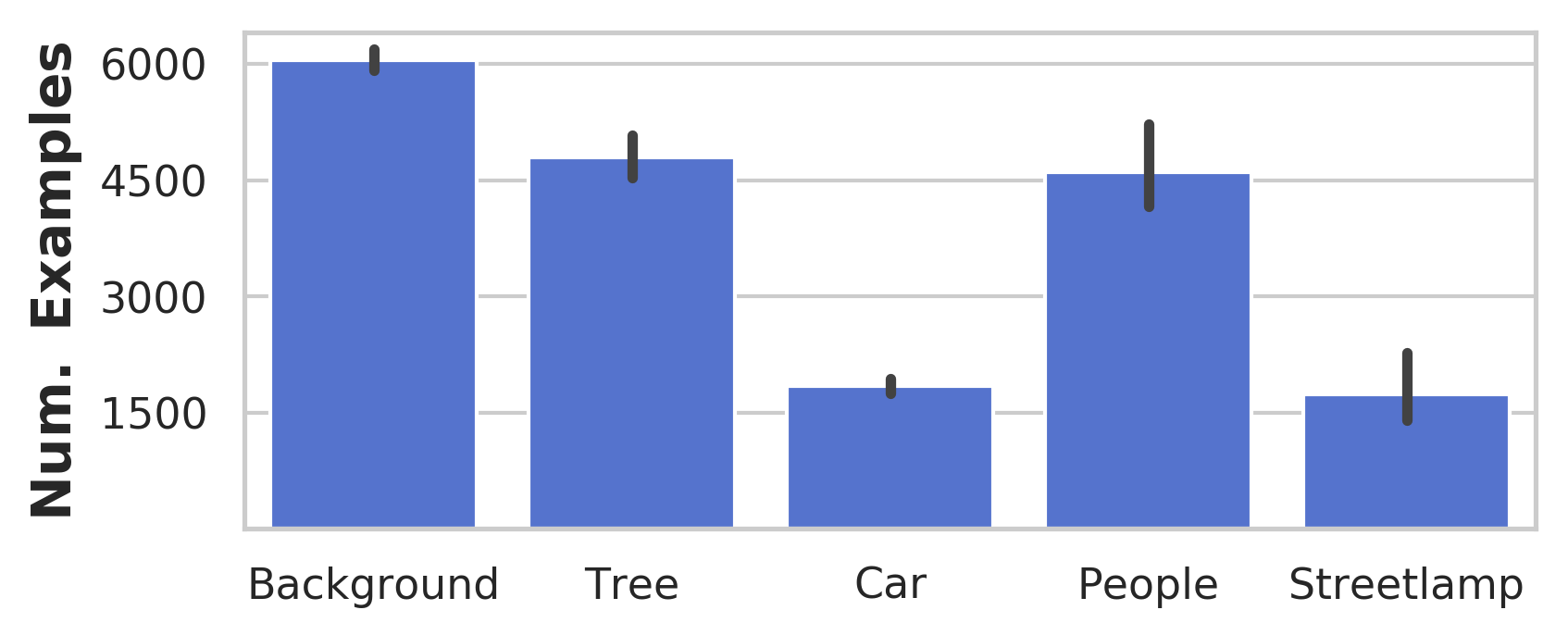}
	\captionof{figure}{Number of image patches per object category extracted from a video sub-sequence according to the procedure presented in supplementary part \ref{section:data_generation}. \label{figure:num_examples}}
\end{wrapfigure}

For our specific continual learning experiments the data generation for all scenario's corresponding five training and testing datasets has been conducted using an overall amount of  $N_S = 150$ street tiles per video sub-sequence. 
Because our used investigated deep convolutional neural network does not explicitly encode temporal dependency between video frames and the extracted image patches no longer contain most of the respective scene context, we have further lowered the capturing rate to 5 frames per second, equalling approximately 15 minutes of discretized video per sub-sequence. Depending on the precisely generated video sequence, the above process configuration yields an approximate amount of $\sim 20000$ classification image patches per video sub-sequence. We show an example illustration of the obtained class balance in figure \ref{figure:num_examples} for one such sub-sequence. 
We can see that the classes background, tree and humans are balanced almost equally, with a factor less than four in terms of decreased amount of overall car and street lamp instances. However, we did not observe any difficulties with respect to training due to this imbalance. 
We re-emphasize the main body's statement that all generated dataset sequences are simply a result of particular choices for the specific continual learning experiments to illustrate their volatility across considered scenarios, but are made publicly available for full reproducibility of exact experiments. The respective detailed set-up for how the constructed classification datasets are used for convolutional neural network training are provided in the upcoming supplementary material section. We specifically encourage future researchers to lift our currently included dataset construction simplifications and evaluate our simulator and respective continual learning techniques in light of more challenging tasks. 

\section{Deep Neural Network Training Hyper-parameters} \label{section:hyper_parameters}
\begin{figure}[h]
	\centering
	\resizebox{0.975\textwidth}{!}{
		\begin{tabular}{p{2.5em} | c c c}
			\hspace*{1em}$\pmb{\pi_t}$ & \textbf{Class Incremental} & \textbf{Light Incremental} & \textbf{Weather Incremental}\\
			\hline \\[-0.3em]
			\vspace*{-4em}$t=1$&
			\includegraphics[width=0.3\textwidth]{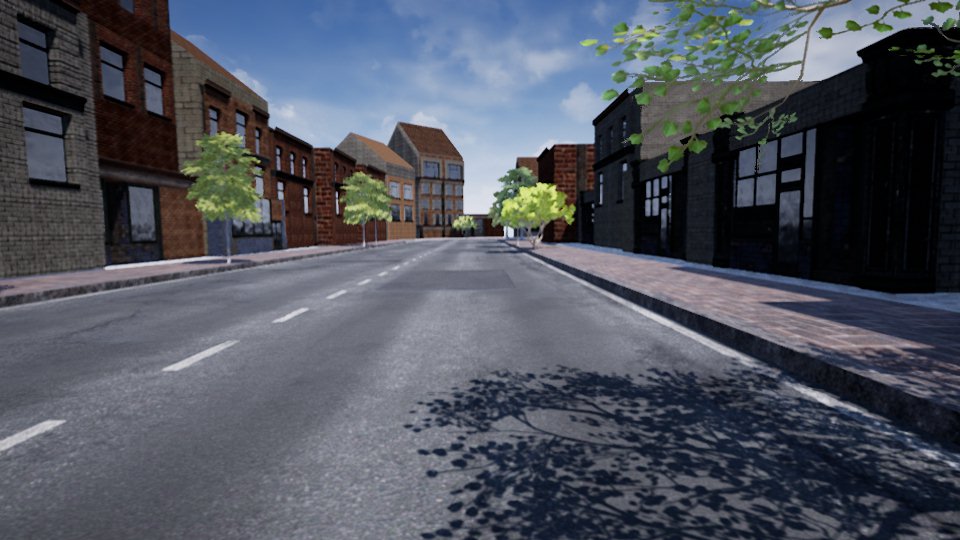} & \includegraphics[width=0.3\textwidth]{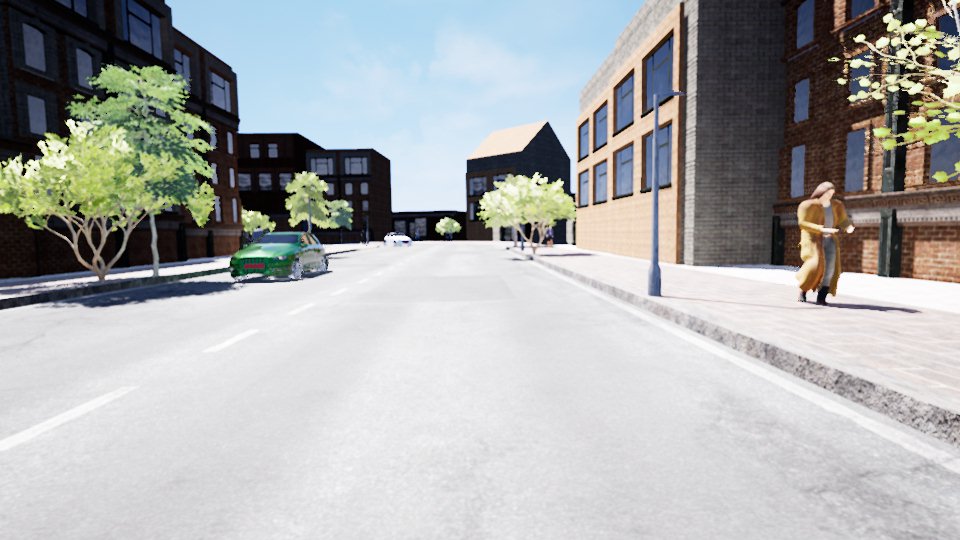} &
			\includegraphics[width=0.3\textwidth]{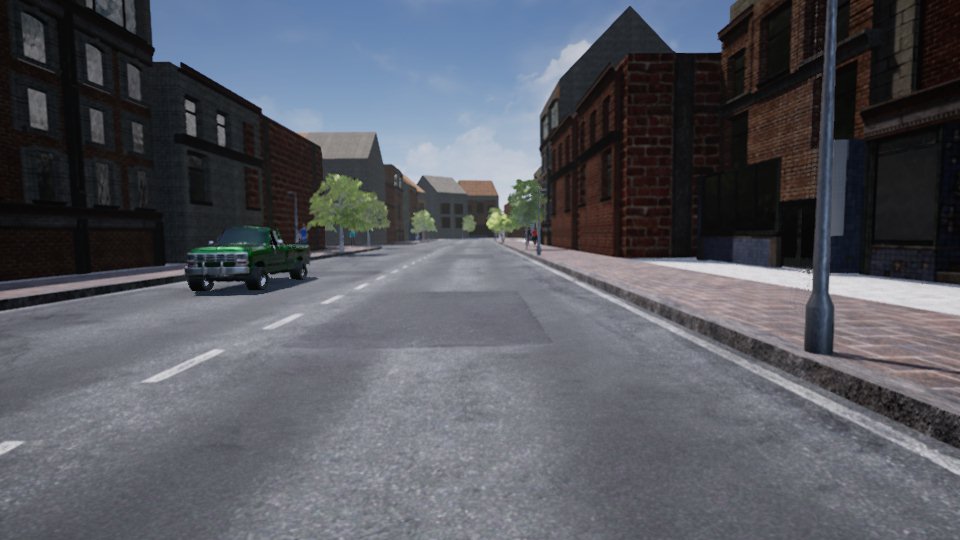}\\
			\vspace*{-4em}$t=2$&
			\includegraphics[width=0.3\textwidth]{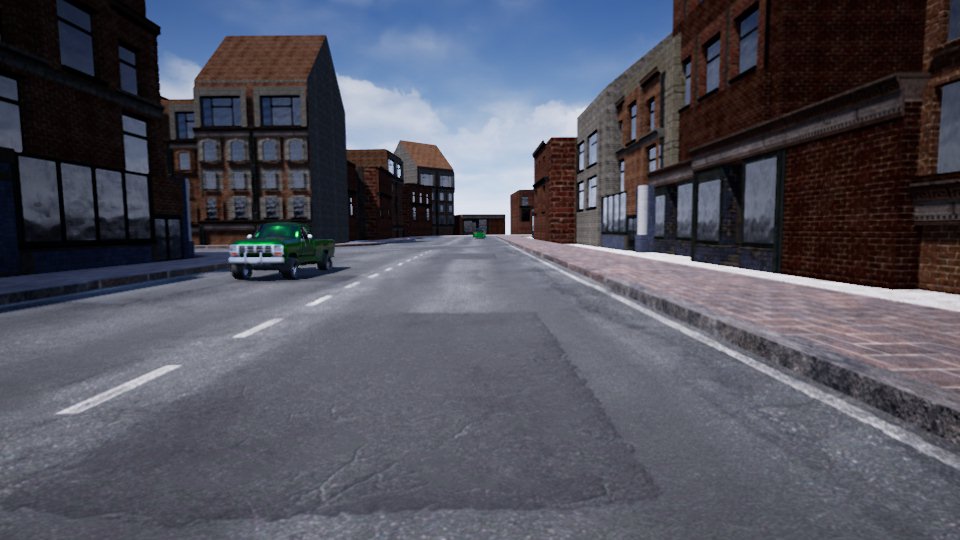} & \includegraphics[width=0.3\textwidth]{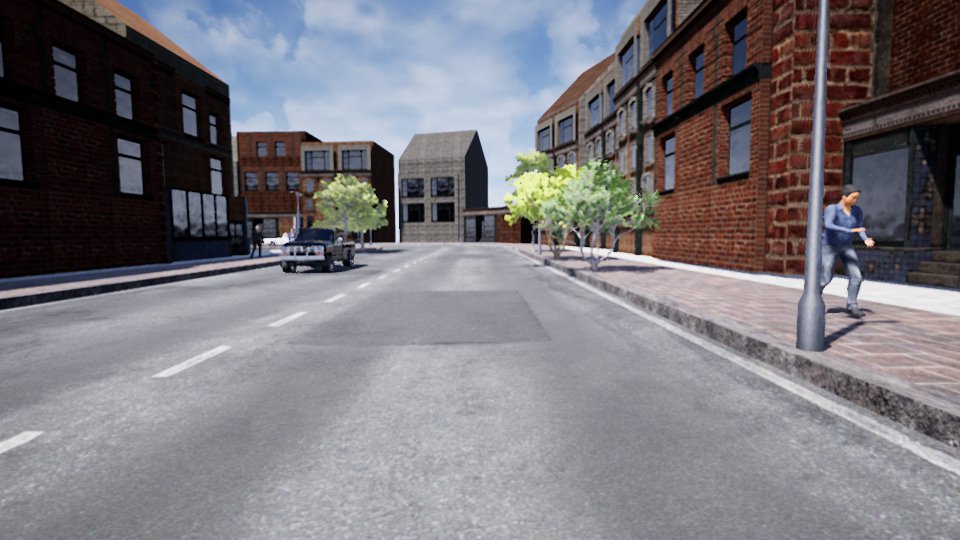} &
			\includegraphics[width=0.3\textwidth]{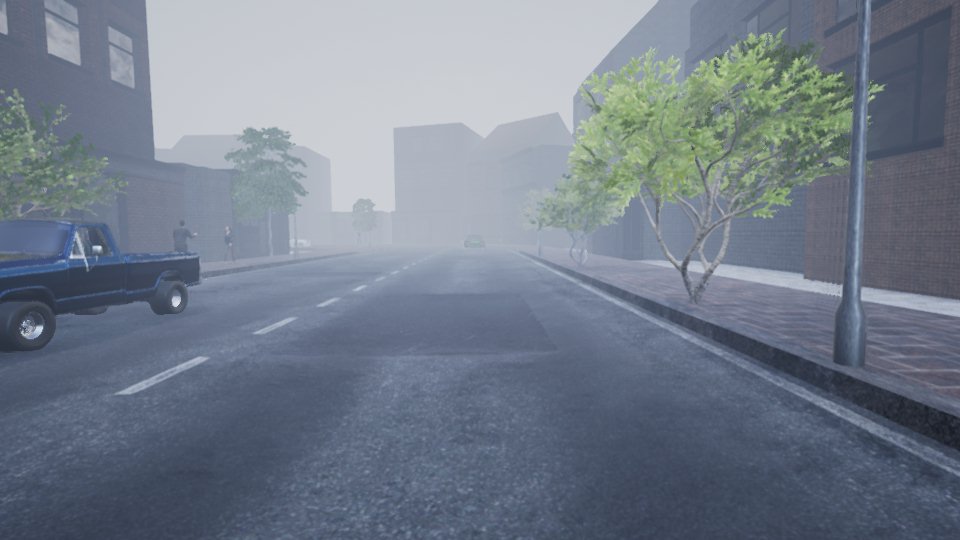}\\
			\vspace*{-4em}$t=3$&
			\includegraphics[width=0.3\textwidth]{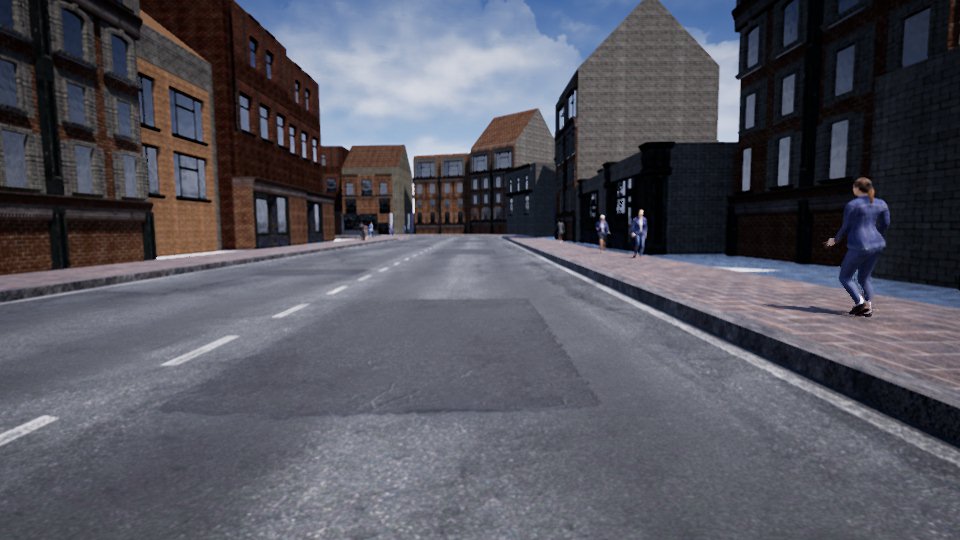} & \includegraphics[width=0.3\textwidth]{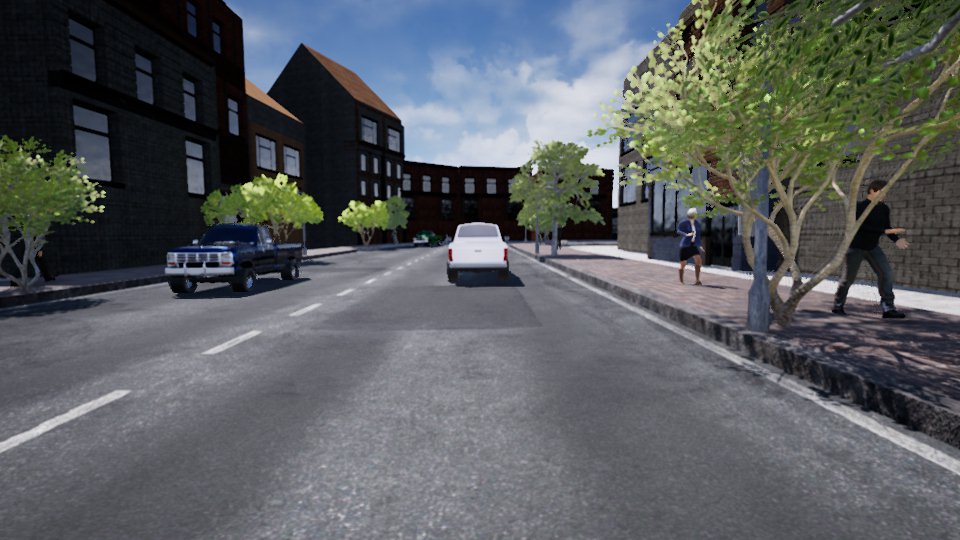} &
			\includegraphics[width=0.3\textwidth]{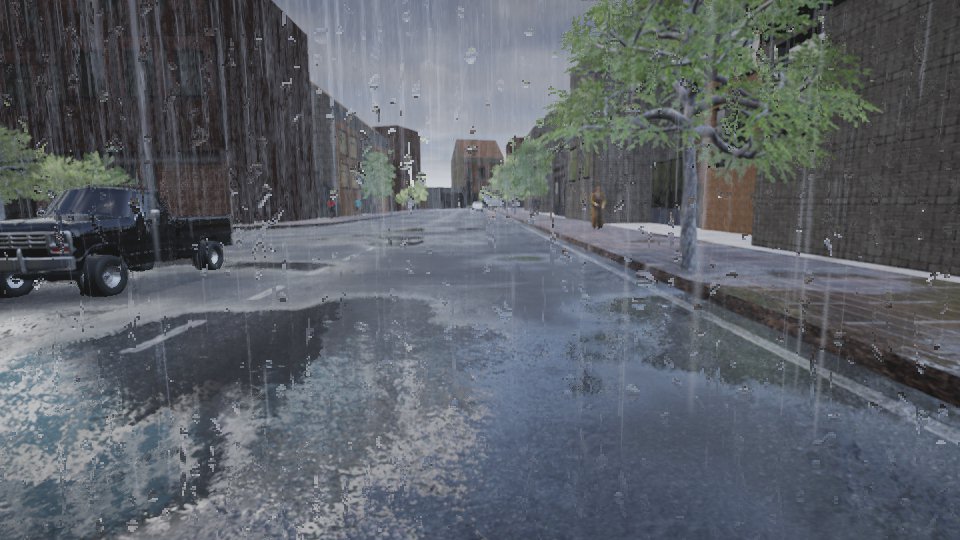}\\
			\vspace*{-4em}$t=4$&
			\includegraphics[width=0.3\textwidth]{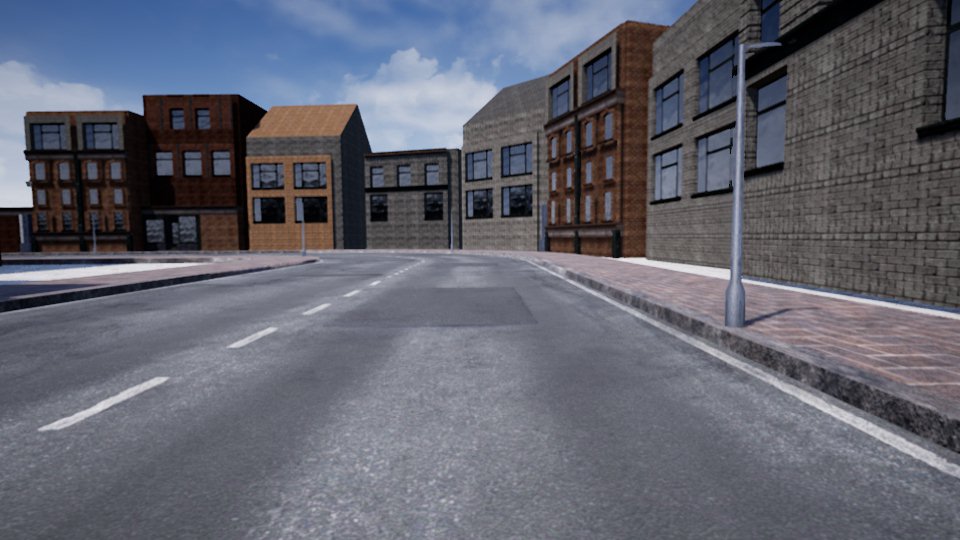} & \includegraphics[width=0.3\textwidth]{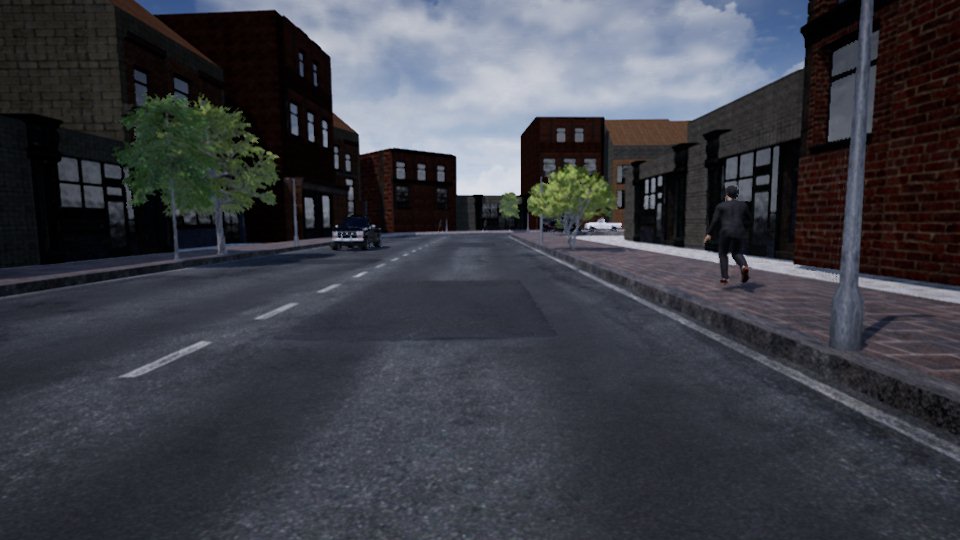} &
			\includegraphics[width=0.3\textwidth]{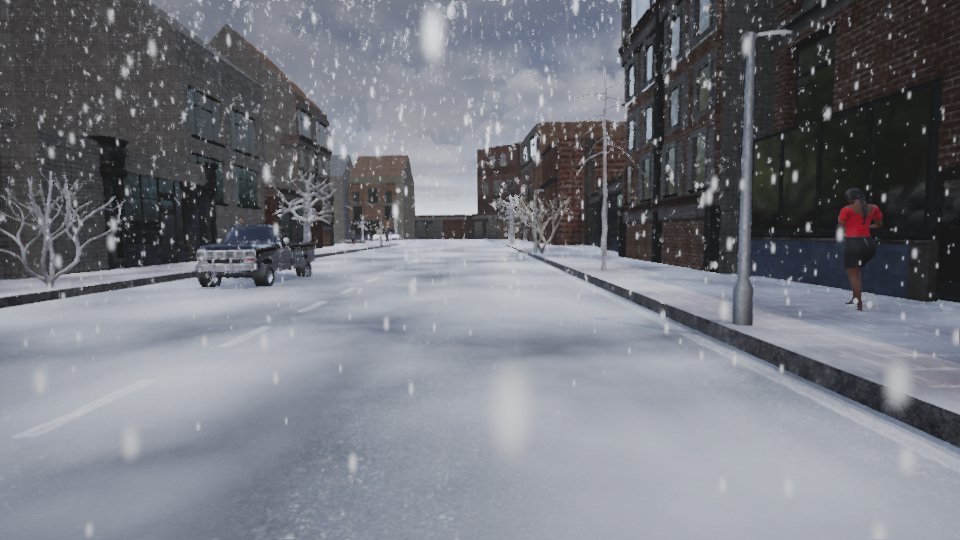}\\
			\vspace*{-4em}$t=5$& & \includegraphics[width=0.3\textwidth]{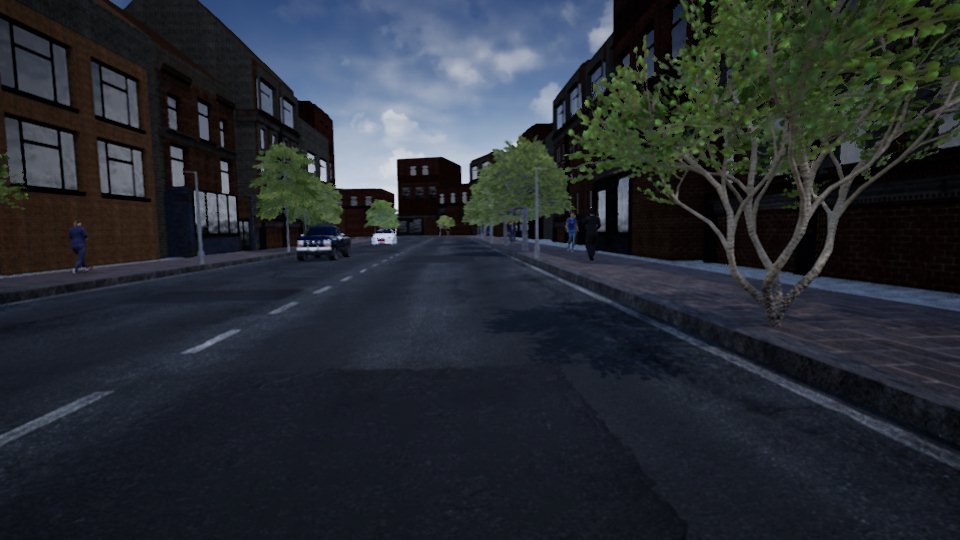} &
			\includegraphics[width=0.3\textwidth]{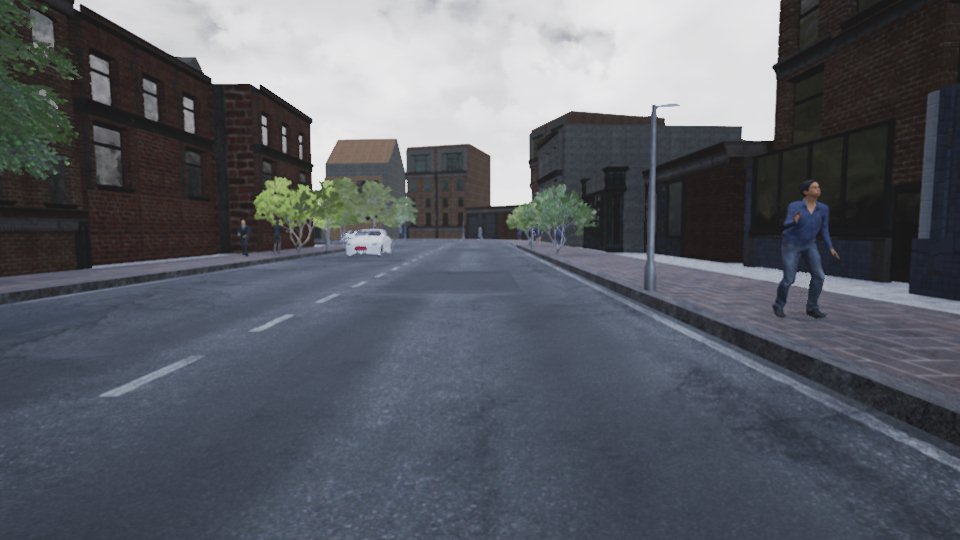}\\
		\end{tabular}
	}
	\captionof{figure}{Snapshot of training sub-sequences for incremental class, lighting and weather, learning scenarios. Recall that the task is always object classification with given location.  \label{figure:snapshot_examples}}
\end{figure}

The preceding supplementary section has provided further details on the main body's experimentation with respect to the constructed and used classification datasets. In this section, we elaborate on the respective hyper-parameters that have been used in conjunction with these datasets. Recall that we have shown empirical results for continual learning classification tasks to contrast five deep continual learning techniques in three scenarios. Consequently the performance of Synaptic Intelligence (SI) \cite{Zenke2017}, Learning without Forgetting (LwF) \cite{Li2016}, Elastic Weight Consolidation (EWC) \cite{Kirkpatrick2017}, Gradient Episodic Memory (GEM) \cite{Lopez-Paz2017}, and Open set Classifying Denoising Variational Auto-Encoders (OCDVAE) \cite{Mundt2020a} has been shown for object classification in incremental scenarios. To reinforce the main body's description of these scenarios in an intuitive form, we show a snapshot of each training video sub-sequence in figure \ref{figure:snapshot_examples}. Here, we can see an illustration of the appearance and disappearance of entire categories in the class incremental scenario (left column), a progressive decrease in lighting intensity (center column), and changes in weather (right column).

To focus on the limitations of our selection of deep continual learning procedures in the paper's main body, even in very simple settings, we have used the previously described constructed image patch datasets and have treated each sub-sequence as a separate dataset. That is, we have trained neural networks for multiple epochs on repeatedly shuffled mini-batches until convergence on one sub-sequence, before proceeding to the next sub-sequence that is then trained analogously. Intuitively, this breaks any assumption of temporal dependency, as our used model does not explicitly encode temporal information and predicts on a patch-by-patch basis. It further lifts any requirement of online learning or prediction. Each video sub-sequence has thus been decomposed into a training dataset that is treated in isolation, as is frequently done in simple computer vision classification benchmarks. The respective generated test sequences have been treated in similar isolation. For each sub-sequence a separate image patch dataset has been constructed. The key difference to the training datasets is that for evaluation the test datasets are always evaluated over all images of all sub-sequences ranging up to the current task, where the training is conducted solely on one task at a time.

To assure that the neural network baseline's accuracy is in principle sufficient to learn our tasks and the generated data is sufficient in amount, we have thus first trained a so called continual learning upper-bound. That is, we have trained on the sequence of training data, whereas all sequential tasks are accumulated similar to the test set to investigate whether we can reach a 100\% accuracy if the neural network has access to all data. This experiment has been validated for each of the three scenarios and has a particular purpose. It allowed an initial investigation with respect to different state-of-the-art neural networks and their basic training hyper-parameters, in which it has quickly been determined that a four layer convolutional architecture based on the popular DCGAN architecture  \cite{Radford2016} suffices for our classification tasks.

The DCGAN based "DCNN" architecture (as we do not train a GAN) serves as a common baseline for all experiments, where the encoder is always trained for all methods, and the mirrored decoder is required exclusively in the variational autoencoder based generative replay experiments. Note that as indicated by the original authors of OCDVAE, the classification nevertheless is based on the latent space resulting from the encoder. 
The main difference is that the generative classifier also takes into account the joint distribution between labels and data, instead of solely predicting the conditional probability of a class given a data point. The precise configuration of encoder and decoder architecture components is shown in tables \ref{table:architecture}. For each convolutional layer, the tables specify the size of the convolutional kernel, the used amount of learnable features, employed zero padding and convolutional strides. Each convolutional layer is to be understood as a block that is followed by batch-normalization \cite{Ioffe2015} with $\epsilon = 10^{-5}$ and rectified linear unit activations. Classifiers are linear in nature and simply project from the resulting feature space of the encoder to a vector of length corresponding to the amount of classes. As the considered classification task is of single target nature, a Softmax function is used to obtain prediction confidences. In case of the variational auto-encoder, this classifier is preceded by the variational re-parametrization.

\begin{table}[h]
	\captionof{table}{4-layer DCNN encoder and decoder architecture. Convolutional (conv) and tranposed convolutional layers (conv\_t) are parameterized by size and number of filters, input padding $p$, and stride $s$.
		Fully-connected layers (fc) are parameterized by their number of input and output units, where $L_\text{Dim}$ is the dimension of the latent space, and $\ Flat_\text{Enc}$ the number of units for the enoder's flattened last layer.
		Each full-connected and convolutional layer is followed by batch-normalization with value of $10^{-5}$, and a rectified linear unit activation function.
		The decoder's ultimate layer ends on a Sigmoid activation function. \label{table:architecture}}
	\vspace*{1em}
	\begin{minipage}{0.5\textwidth}
		\vspace*{-1.05em}
		\hspace*{0.75em}
		\resizebox{0.9\textwidth}{!}{
			\begin{tabular}{p{0.21\textwidth} | p{0.1\textwidth} p{0.52\textwidth}}
				Layer & Encoder\\
				\hline\\[-1em]
				\textit{Encoder} 1 & (conv) & $4\times4,\ 128, \ p=1, \ s=2$ \\
				\textit{Encoder} 2 & (conv) & $4\times4,\ 256, \ p=1, \ s=2$ \\
				\textit{Encoder} 3 & (conv) & $4\times4,\ 512, \ p=1, \ s=2$ \\
				\textit{Encoder} 4 & (conv) & $4\times4,\ 1024, \ p=0, \ s=2$
				\\	
			\end{tabular}
		}
	\end{minipage}
	\begin{minipage}{0.55\textwidth}
		\hspace*{-0.75em}
		\resizebox{0.9\textwidth}{!}{
			\begin{tabular}{p{0.21\textwidth} | p{0.11\textwidth} p{0.5\textwidth}}
				Layer & Decoder\\
				\hline \\[-1em]
				\textit{Decoder} 1 & (fc) & $ L_\text{Dim} \rightarrow  Flat_\text{Enc}$ + Reshape\\
				\textit{Decoder} 2 & (conv\_t) & $4\times 4 - 512 \ p=0, \ s=2$\\
				\textit{Decoder} 3 & (conv\_t) & $4\times 4 - 256 \ p=1, \ s=2$\\
				\textit{Decoder} 4 & (conv\_t) & $4\times 4 - 128 \ p=1, \ s=2$\\
				\textit{Decoder} 5 & (conv) & $4\times 4 - 3 \ p=1, \ s=1$ \\
			\end{tabular}
		}
	\end{minipage}
\end{table}
An adequate length of the training procedure on shuffled mini-batches of the extracted patch dataset for each video sub-sequence has empirically been determined to correspond to $60$ epochs for encoder only models and $120$ epochs for VAE models, after which convergence of losses has been observed. Optimization is conducted using an Adam optimizer, with an identified learning-rate of $0.001$, and the momenta parameter set to $\beta = (0.9, 0.999)$, as proposed by Kingma et al. \cite{Kingma2015}. All model weights are initialized according to He et al. \cite{He2015}. The used mini-batch size is $64$, and all quadratic image patches have been resized to meet a static input resolution of $64 \times 64$, with no further data augmentation applied. Realization of the experimentation has been conducted in approximately $800$ GPU hours on an NVIDIA A100-SXM4-40GB GPU cluster using the repository provided in the main body, which is a fork of the public OCDVAE codebase \cite{Mundt2020a} in combination with the Avalanche \cite{Lomonaco2021} continual learning library.

Apart from these general shared training hyper-parameters, the five deep continual learning methods each come with their own additional hyper-parameters. For the reader's convenience we also provide a short summary of each method, to provide a better intuition behind the hyper-parameters' significance. 

\begin{itemize}
	\item \textbf{LwF:} Learning without Forgetting  \cite{Li2016} is a functional regularization mechanism that relies on knowledge distillation \cite{Hinton2014}. Based on the latter, a trained classifier for an initial task is used as a reference for desired predictions of the task's classes, so called soft-labels, when proceeding to train the classifier on additional classes of subsequent tasks. Specifically, before updating the encoder's and classifier's weights to incorporate new classes, their output vector is recorded. Even though the new class labels are not yet included, the assumption is that preserving this output to an extent also prevents forgetting of the older classes for data instances where the prediction is actually correct. As such, a hyper-parameter $\lambda_{LwF}$ controls the strength with which the additional loss term that imposes this constraint acts on the continual learning. 
	\item \textbf{EWC and SI:} Elastic Weight Consolidation \cite{Kirkpatrick2017} and Synaptic Intelligence \cite{Zenke2017} both employ parameter regularization by associating each learned parameter with an importance measure. 
	In EWC this measure is based on the diagonal of the Fisher information matrix for each weight layer. For SI it is derived from the extent to which specific parameters have contributed to the total loss decrease. Assuming a generally over-parametrized deep neural network, an additional quadratic loss is then imposed in order to prevent changes to the subset of important parameters. The strength of the loss that simply minimizes the difference between the previous important parameter state and the next is controlled by a hyper-parameter $\lambda_{EWC}$ and $\lambda_{SI}$ respectively.
	\item \textbf{GEM:} Gradient Episodic Memory \cite{Lopez-Paz2017} is a hybrid approach of regularization that includes an additional memory buffer to store examples for each learned task. Stored samples are used to impose constraints on gradient updates for future task learning. During optimization, the gradient obtained from the observed new examples is projected using gradients recovered from each task memory, such that the resulting parameter update of the model shall not increase the loss on any of the memorized examples of a previous task. A respective hyper-parameter $\lambda_{GEM}$ (gamma in the original paper) biases the gradients' projection to favor backwards transfer, as stated by the authors. 
	\item \textbf{Exemplar Replay:} 
	Exemplar replay utilizes an external memory and a pattern storing heuristic to interleave the model's training process for a continual learning mechanism \cite{Rebuffi2017, Nguyen2018, Bachem2015, Prabhu2020}. In our implementation we use a naive heuristic where for each trained task a number $\epsilon$ of randomly selected patterns from the experienced data is stored to the external memory. During training for any new task, the resulting mini-batches are interleaved with patterns from the external memory, such that they are balanced to be comprised from patterns of each experience.
	\item \textbf{OCDVAE:} Open set Classifying Denoising Variational Auto-Encoder \cite{Mundt2020a} relies on training a deep generative model to replay data in order to prevent older tasks from being forgotten. It uses the VAE's approximation to the data distribution and employs an open set recognition mechanism to sample data distribution inliers for formerly seen tasks and decodes these sampled values into images to rehearse. A respective hyper-parameter serves as a constraint on the inlier likelihood that is expected to yield clear and representative exemplars. 
\end{itemize}

Following the original authors' suggestions we assume a 60-dimensional latent space for OCDVAE, use a unit Gaussian prior and consider sampled data points with an outlier likelihood of less than 5\% to be key representatives of the dataset. With respect to SI and LwF, we have observed the choice of $\lambda_{LwF}$ and $\lambda_{SI}$ to barely influence the experimental outcome. In our main body's experiments we have used values of $\lambda_{SI} = \lambda_{LwF} = 0.5$.
These parameters have been verified by a preliminary grid-search which revealed no fundamental deviations in performance for $\lambda$-values of $0.1$, $0.5$, $1.0$, $10.0$, and $100.0$. Regarding the sizes for external memory of GEM and the Exemplar Replay approach, we allowed storing of $\epsilon = 200$ examples per task, which amounts to approximately $1\%$ to $4\%$ of the total dataset's size of the respective scenarios. 

\section{Operators for Quasi-invariance to Illumination} \label{section:invariant_transforms}
In the main body we have emphasized the lack of robustness of some continual learning approaches, particularly in scenarios such as the incremental lighting setting, where the overall illumination intensity is varied. For the latter, we have shown that rather simple transformations, as a pre-processing step to operate on a space that is quasi-invariant to homogeneous illumination changes, can solve the imposed continual learning task, without the necessity of any additional mechanism to prevent catastrophic forgetting. Naturally, this is due to the image not changing in the quasi-invariant space when the illumination intensity is varied, much in contrast to the original raw RGB color image input that a deep neural network is typically expected to process.

\begin{figure}[b]
	\newcommand{\x}{0.12\textwidth}
	\newcommand{\y}{0.11\textwidth}
	\centering
	\begin{tabular}{p{0.05\textwidth} p{\y} p{\y} p{\y} p{\y} p{\y}}
		\vspace*{-3em}\textbf{RGB}& \includegraphics[width=\x]{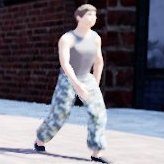} &
		\includegraphics[width=\x]{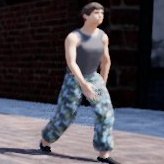} & 
		\includegraphics[width=\x]{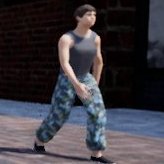}& 
		\includegraphics[width=\x]{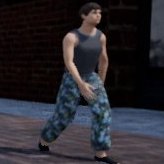}&
		\includegraphics[width=\x]{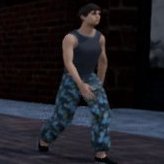}\\
		\\[-0.8em]
		\vspace*{-3em}\textbf{LBP}&
		\includegraphics[width=\x]{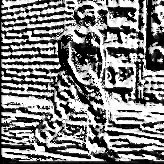} &
		\includegraphics[width=\x]{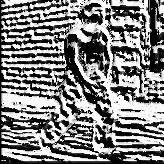} & 
		\includegraphics[width=\x]{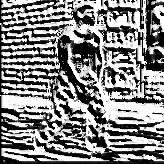}& 
		\includegraphics[width=\x]{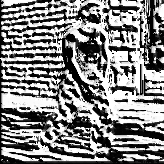}&
		\includegraphics[width=\x]{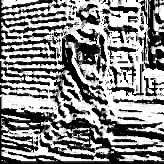}\\
	\end{tabular}
	\captionof{figure}{Illustration of the LBP transform applied to the incremental lighting scenario for the example of a pedestrian. \label{figure:LBP}}
\end{figure}

\begin{figure}[h]
	\newcommand{\x}{0.12\textwidth}
	\newcommand{\y}{0.11\textwidth}
	\centering
	\begin{tabular}{p{\y} p{0em} p{\y} p{\y} p{\y}}
		\hspace*{1.5em}\textbf{RGB} & & \hspace*{2.2em}\large$\pmb{c_1}$ & \hspace*{2.2em}\large$\pmb{c_2}$ & \hspace*{1.8em}\large$\pmb{c_3}$\\
		\hline
		\\[-0.5em]
		\includegraphics[width=\x]{gfx/Example_LightInvaraint/cuts/78.jpeg} & &
		\includegraphics[width=\x]{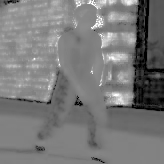} &
		\includegraphics[width=\x]{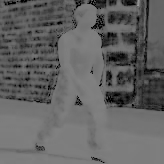} &
		\includegraphics[width=\x]{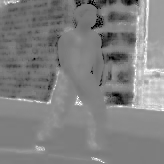}
		\\
		\includegraphics[width=\x]{gfx/Example_LightInvaraint/cuts/19.jpeg}& &
		\includegraphics[width=\x]{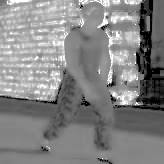} &
		\includegraphics[width=\x]{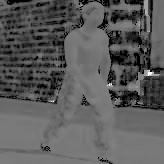} &
		\includegraphics[width=\x]{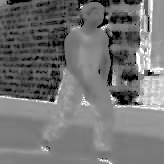}
		\\
		\includegraphics[width=\x]{gfx/Example_LightInvaraint/cuts/9_6.jpeg}& &
		\includegraphics[width=\x]{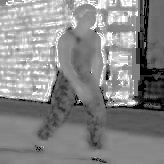} &
		\includegraphics[width=\x]{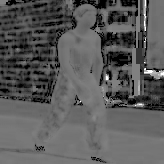} &
		\includegraphics[width=\x]{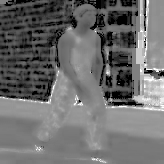}
		\\
		\includegraphics[width=\x]{gfx/Example_LightInvaraint/cuts/2_4.jpeg} & &
		\includegraphics[width=\x]{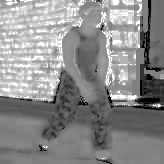} &
		\includegraphics[width=\x]{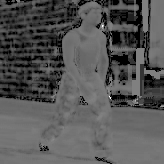} &
		\includegraphics[width=\x]{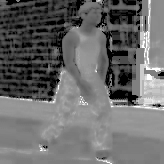}
		\\
		\includegraphics[width=\x]{gfx/Example_LightInvaraint/cuts/1_2.jpeg} & & \includegraphics[width=\x]{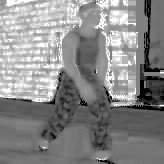} &
		\includegraphics[width=\x]{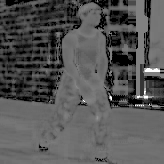} &
		\includegraphics[width=\x]{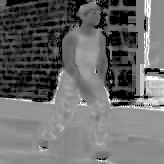} 
		\\
	\end{tabular}
	\captionof{figure}{Illustration of the color ratio transform applied to the incremental lighting scenario for the example of a pedestrian. \label{figure:color-ratio}}
\end{figure}

To provide the visual intuition behind the lighting invariant transformations, we provide illustrations of snapshots showcasing a simulated human with varied degrees of lighting strength and the respective image after transformation in figures \ref{figure:LBP} and \ref{figure:color-ratio}. Here, figure \ref{figure:LBP} exhibits the quasi-invariance of the space resulting from application of local binary patterns (LBP) \cite{He1990,Wang1990}. Figure \ref{figure:color-ratio} provides an analogous depection for the raw image transformation through calculation of color ratios \cite{Gevers1999}, which yield a similarly invariant space under the assumption of non-changing illumination color. The chosen illumination intensity in these figures corresponds to the precise Lux values assumed in the main body's experiments. To explain how these visualized spaces are formed, we briefly summarize the core concepts of the two respectively investigated transformations. 

The quasi illumination invariance of the LBP operator originates from an encoding of pixels according to their relative relations to their neighbors in a limited spatial vicinity. That is, within a chosen radius the operation checks whether values have greater or smaller magnitude. In the case of homogeneous lighting variations, these relationships remain unaltered. For our specific experiments of the main body, the LBP radius has been set to $3$, i.e. including 24 neighboring pixels. We note that we have found no necessity to tune this value and have refrained from investigation of different radii, as the initially chosen radius immediately resulted in almost perfect accuracy in the main body's experiment. This is not very surprising as the transformed images of figure \ref{figure:LBP} are almost identical and learning the task on the any sub-sequence is expected to provide the solution to all other sub-sequences. 

The quasi illumination invariance of Gever's and Smeulder's color constancy transformation results from the simple insight that ratios of individual intensity channels remain constant under varying lighting intensity. Recalling the transformation for an individual image channel as introduced in the main body: $c_1 = \arctan (R / \mathtt{max} \{ G, B\} )$, and corresponding definitions for the other two channels, we can qualitatively grasp this concept for the images presented in figure \ref{figure:color-ratio}. Note how the sequence of raw RGB images and their transformation also nicely illustrates the quasi-invariance, in contrast to an ideally desired full invariance. In the case of too strong illumination intensity coupled with a specific set of non-adaptive camera parameters, we can observe how the brightest image suffers from an onset of over-exposure. As a result, the transformed image features some discrepancy to the other transformed images that remain unchanged for the other considered intensities. 
Instinctively, this explains why the resulting final accuracy of this approach in the main body is worse than that of LBP, although we reinforce that it still provides significant improvement over processing of raw images. 

The short visual examples of this section and their respective quantitative results presented in the main body reinforce our message that continual learning can benefit from further explicit modelling beyond relying on representations derived exclusively from raw data. In particular, operation in quasi-invariant spaces can provide a massive benefit and sometimes a straightforward solution to the catastrophic forgetting challenge. We have chosen the two above operators to illustrate the importance of taking into account the application context and encourage future researchers to conduct additional investigations towards a symbiosis of quasi-invariant operators and deep neural networks with the help of our proposed continual simulation framework. 
	
\end{document}